\documentclass[Afour, sageh, times, doublespace]{sagej}

\usepackage{moreverb, url}
\usepackage[colorlinks,bookmarksopen,bookmarksnumbered,citecolor=red,urlcolor=red]{hyperref}

\newcommand\BibTeX{{\rmfamily B\kern-.05em \textsc{i\kern-.025em b}\kern-.08em
T\kern-.1667em\lower.7ex\hbox{E}\kern-.125emX}}

\usepackage{hyperref}
\usepackage{siunitx}
\DeclareSIUnit{\rad}{rad}
\usepackage[justification=centering]{subfig}
\usepackage{import}

\DeclareSIUnit\rpm{rpm}

\usepackage{multirow}
\usepackage{graphicx}

\begin{document}

\runninghead{The Rosario Dataset v2}

\title{The Rosario Dataset v2: Multimodal Dataset for Agricultural Robotics}

\author{Nicolás Soncini\affilnum{1,*}, Javier Cremona\affilnum{1,*}, Erica Vidal\affilnum{1}, Maximiliano García\affilnum{1}, Gastón Castro\affilnum{2}, Taihú Pire\affilnum{1}}

\affiliation{\affilnum{1}CIFASIS (CONICET-UNR), Rosario, Santa Fe, Argentina\\
\affilnum{2}Universidad de San Andr{\'e}s (UDESA-CONICET), Buenos Aires, Argentina\\
\affilnum{*}Equally contributing authors}
\corrauth{Nicolás Soncini, CIFASIS (CONICET-UNR), Bv. 27 de Febrero 210 bis, S2000EZP, Rosario, Santa Fe, Argentina}
\email{soncini@cifasis-conicet.gov.ar}

\begin{abstract}
We present a multi-modal dataset collected in a soybean crop field, comprising over two hours of recorded data from sensors such as stereo infrared camera, color camera, accelerometer, gyroscope, magnetometer, GNSS (Single Point Positioning, Real-Time Kinematic and Post-Processed Kinematic), and wheel odometry. 
This dataset captures key challenges inherent to robotics in agricultural environments, including variations in natural lighting, motion blur, rough terrain, and long, perceptually aliased sequences. 
By addressing these complexities, the dataset aims to support the development and benchmarking of advanced algorithms for localization, mapping, perception, and navigation in agricultural robotics.
The platform and data collection system is designed to meet the key requirements for evaluating multi-modal SLAM systems, including hardware synchronization of sensors, 6-DOF ground truth and loops on long trajectories.

We run multimodal state-of-the art SLAM methods on the dataset, showcasing the existing limitations in their application on agricultural settings.
The dataset and utilities to work with it are released on \url{https://cifasis.github.io/rosariov2/}.

\keywords{dataset, multi-modal, agriculture, crop field, ground-truth, soybean field, unstructured}
\end{abstract}

\maketitle

\section{Introduction} \label{sec:introduction}

World population will grow by a third by 2050, directly impacting global food demand (\cite{fukase2020economic}). In this context, the agricultural industry should increase its production to satisfy such demand. The use of autonomous robots to carry out agricultural tasks such as seeding, harvesting,  weed remotion, pest control among others is an attractive solution since it can improve the production time in a sustainable manner reducing the environmental impact and pollution.

\begin{figure}[!htbp]
    \centering
    \includegraphics[width=\linewidth]{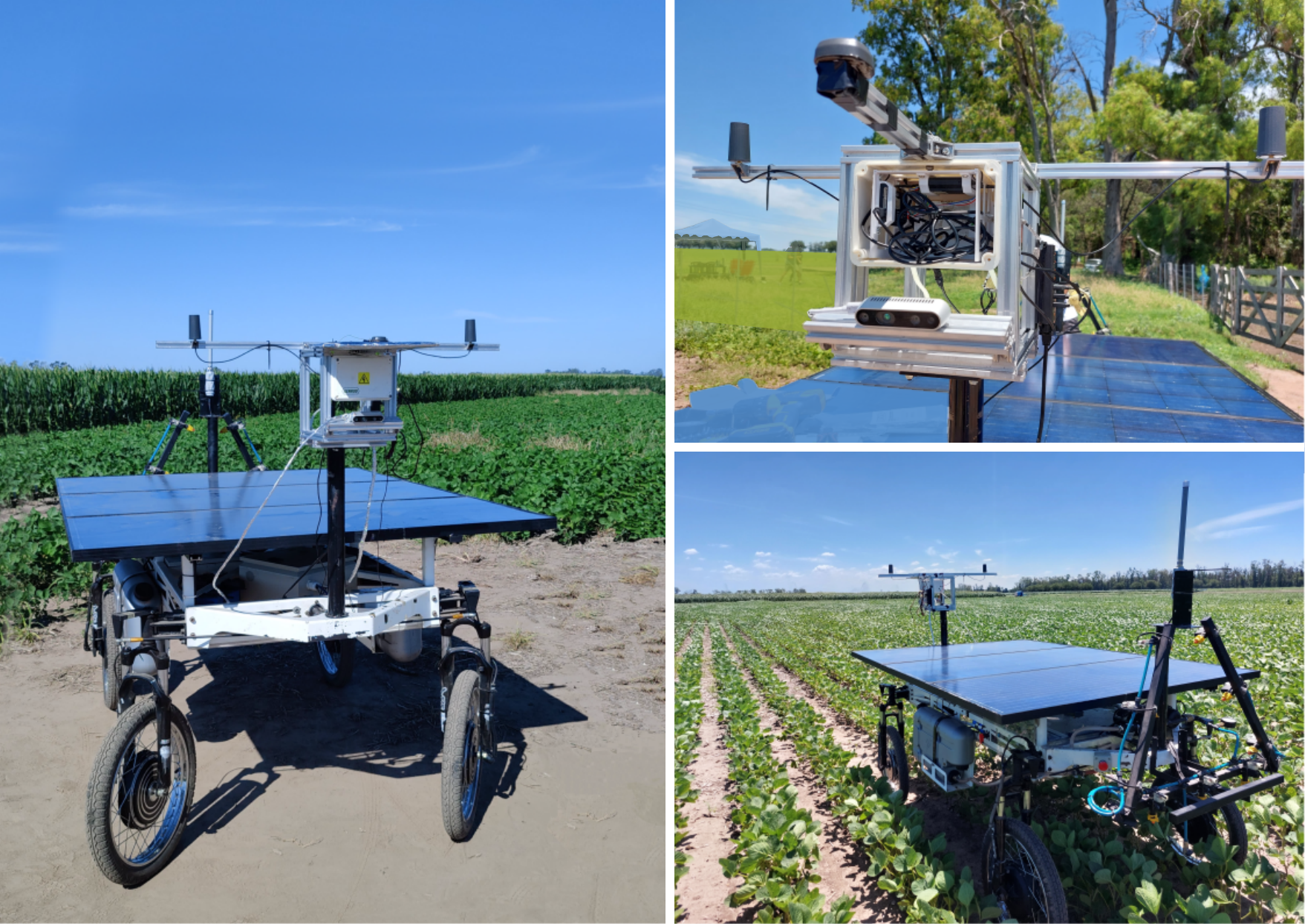}
    \caption{Photos of the weed removing robot on the field and a close-up of the sensor suite it carries on its front.}
    \label{fig:weed_removing_robot}
\end{figure}

However, the implementation of autonomous robots in the agricultural field is a challenging work due the rough terrain, natural light variations, perceptual aliasing, areas with GNSS-denied signal, and the long-term robot operation required to carry out the desired applications. All these challenges cause robot localization methods to fail or perform poorly, making them impractical for real agricultural tasks, as evidenced in  \cite{cremona2022evaluation,cremona2023gnss,soncini2024addressing};
\cite{cox2023, bai2023};
\cite{ait2023travelling}.

In the last decade, there has been a growing trend towards the creation and public availability of agricultural datasets, enabling researchers to test new techniques and develop more sophisticated algorithms to address these challenges, such as \cite{pire2019rosario,kragh2017fieldsafe, marzoatanco2024magrodataset}. 
However, none of them are properly curated for evaluating multi-modal SLAM algorithms.
Effective multi-modal SLAM evaluation imposes specific requirements such as hardware-synchronized sensors, 6-DOF ground-truth, and trajectories with loops to effectively test loop closure algorithms. 

In this work, we present a multi-modal dataset recorded by the weed removing robot developed at CIFASIS (CONICET-UNR) in a soybean agricultural field.
The recorded data is captured by a RGB camera, stereo IR camera, a 6-Axis IMU, three 9-Axis IMU, three GNSS receivers and wheel encoders, and we provide a synchronization system for the data produced.
The dataset consists of six sequences where the robots covers a total of \SI{7.3}{\kilo\meter} in \SI{2.2}{\hour}.
The recorded data captured natural light changes, image blurred, long trajectories, loop closures, rough terrain navigation.
In order to illustrate the dataset challenges, we evaluate and analyze multi-modal state of the art SLAM systems on it.
The dataset was conceived to facilitate the development of new multi-modal SLAM algorithms for agricultural environments.
We also provide a 6-DoF ground-truth obtained by two GNSS and inertial measurements. 

The main contributions of this works can be summarized in the following:
\begin{itemize}
    \item Multi-sensor data captured by RGB camera, stereo IR camera, a 6-Axis IMU, three 9-Axis IMU, three GNSS receivers and wheel encoders.
    \item 6-DoF pose ground-truth generated by two GNSS and inertial measurements.
    \item Trajectories with loops, which are trajectories where the robot revisits the same location after traveling along a certain path.
    \item A set of tools used to handle the data, provided as open-source.
    \item Sensor calibration sequences.
    \item Data is provided in both ROS rosbags and plain files.
\end{itemize}

The dataset and the development tools are release as open-source to facilitate the data usage and the development of new SLAM multi-modal systems.

We organize the rest of the article as follows: in Section~\ref{sec:related} we present the most recent and relevant work related to datasets for agricultural robotics. In Section~\ref{sec:platform} we describe the physical setup of the robot used to record the present dataset as well as the software and the calibration methodology applied.
In Section~\ref{sec:dataset} we showcase the recorded data released in the dataset, along with information about the recording methodology and pre and post-processing of the data.
In Section~\ref{sec:results} we analyze results of running state-of-the-art algorithms for localization, mapping and 3D reconstruction on the data in an attempt to prove its worthiness and show what still remains to be addressed for enhancing these methods on agricultural environments.
Finally, Section~\ref{sec:conclusions} summarizes the presented article and discusses future work.

\section{Related Work} \label{sec:related}

In the field of agriculture there are many types of crops, terrains and farming practices, each creating different visual imprints and navigation patterns.
Our work continues what started in \cite{pire2019rosario} as it uses the same robot platform and navigates similar conditions (a soybean plantation in the form of rows), but we improve the quality and diversity of sensors, and address long-term trajectories with loops. 
Particularly, we have replaced the Stereolabs ZED camera, consisting of a stereo pair of rolling-shutter color cameras, for an Intel D435i RGB-D, comprising a stereo pair of infrared (IR) global-shutter cameras, a rolling-shutter color camera and an inertial measurement (IMU) unit.
A set of double-band (L1/L2) GNSS modules was added as well, to enhance the positioning for ground-truth calculation and provide an estimate of platform orientation, which was missing in the previous dataset.
Improvements were made to the sensor synchronization as well, which we elaborate on in coming sections.

In Table~\ref{tab:dataset_survey}, we survey the most recent and relevant publicly available datasets focused on robot navigation and localization in agricultural environments. In order to put the proposed dataset in context of the state of the art, we analyzed each dataset according the type of environment, sensors captured, software/hardware synchronization and ground-truth type.
Datasets like Sugar Beets (\cite{chebrolu2017sugarbeet}), Ladybird Cobbity Brassica (\cite{bender2020ladybirdsbrassica}) and RumexWeeds (\cite{guldenring2023rumexweeds}) provide only downwards facing cameras, which makes them very convenient for weed detection and classification tasks, but very poor in terms of localization and mapping.
Others like FieldSAFE (\cite{kragh2017fieldsafe}) and The GREENBOT Dataset (\cite{canadasaranega2024greenbotdataset}) contain very good multisensor data on the field, but provide no odometry information, which can be useful to compensate for IMU drift or continue the position tracking when other sensors fail to do so.
Some notable datasets like Bacchus Long-Term (\cite{polvara2023bacchuslongtermblt}), MAgro (\cite{polvara2023bacchuslongtermblt}), BotanicGarden (\cite{liu2024BotanicGarden}), Terrasenta Under-Canopy (\cite{cuaran2024terrasentiaundercanopy}) and ARD-VO (\cite{crocetti2023ardvo}) also contain high quality multimodal data in agricultural settings, but work in settings where the crops or plants raise higher than their sensors, which present a different set of problems than working over the crops as they form corridor-like structures around the robot. 
Another dataset to mention is WeedMap (\cite{sa2018weedmap}), which captures all data from a multispectral camera mounted on a drone to generate composite images or orthomosaics and allows for crop analysis from this data.
In our dataset we attempt to cover most of the weak points we mention, to provide a rich and heterogeneous set of data to further the research on these environments.

\begin{table*}
\centering
\resizebox{\textwidth}{!}{%
\begin{tabular}{cccccccccc}
\hline
    Dataset & Purpose & Environment & Crop Type & Cameras & IMU & GNSS & Odometry & \begin{tabular}[c]{@{}c@{}}Ground\\ -Truth\end{tabular} & Synchronization  \\ \hline
    \begin{tabular}[c]{@{}c@{}}FieldSAFE\\ \cite{kragh2017fieldsafe}\end{tabular} & \begin{tabular}[c]{@{}c@{}}Obstacle\\ detection\end{tabular} & \begin{tabular}[c]{@{}c@{}}Outdoors;\\ Over crop;\\ Grass\end{tabular} & Grass &  \begin{tabular}[c]{@{}c@{}}Stereo RGB;\\ 360 RGB;\\ Thermal\end{tabular}       & Yes & \begin{tabular}[c]{@{}c@{}}IMU-Fused\\ RTK\end{tabular}          & No       & \begin{tabular}[c]{@{}c@{}}Position;\\ Orientation\end{tabular}  & \begin{tabular}[c]{@{}c@{}}Hardware +\\ Softwre\end{tabular} \\ \hline
    \begin{tabular}[c]{@{}c@{}}Sugar Beets\\ \cite{chebrolu2017sugarbeet}\end{tabular} & \begin{tabular}[c]{@{}c@{}}Weed Detection;\\ Localization;\\ Mapping\end{tabular} &  \begin{tabular}[c]{@{}c@{}}Outdoors;\\ Over crop;\\ Crop rows\end{tabular} & Sugar Beet & \begin{tabular}[c]{@{}c@{}}RGB;\\ IR Depth;\\ Multispectral\end{tabular}       & No  & PPP; RTK & Yes      & Position  & Software \\ \hline
    \begin{tabular}[c]{@{}c@{}}WeedMap\\ \cite{sa2018weedmap} \end{tabular} & \begin{tabular}[c]{@{}c@{}}Weed Detection;\\ Mapping\end{tabular} &  \begin{tabular}[c]{@{}c@{}}Outdoors;\\ Over crop;\\ Crop rows\end{tabular} & Sugar Beet &  Multispectral & Yes & Single-Point & No & Position & - \\ \hline
    \begin{tabular}[c]{@{}c@{}}The Rosario Dataset\\ \cite{pire2019rosario}\end{tabular} & \begin{tabular}[c]{@{}c@{}}Localization;\\ Mapping\end{tabular} &  \begin{tabular}[c]{@{}c@{}}Outdoors;\\ Over crop;\\ Crop rows\end{tabular} & Soybean & Stereo RGB & Yes & RTK & Yes & Position & Software \\ \hline
    \begin{tabular}[c]{@{}c@{}}Ladybird Cobbity Brassica\\ \cite{bender2020ladybirdsbrassica}\end{tabular} & \begin{tabular}[c]{@{}c@{}}Localization;\\ Mapping;\\ Classification\end{tabular} &  \begin{tabular}[c]{@{}c@{}}Outdoors;\\ Over crop;\\ Crop rows\end{tabular}  & \begin{tabular}[c]{@{}c@{}}Cauliflower;\\ Broccoli\end{tabular} &  \begin{tabular}[c]{@{}c@{}}Stereo RGB;\\ Thermal;\\ Multispectral\end{tabular} & Yes & \begin{tabular}[c]{@{}c@{}}IMU-Fused\\ RTK\end{tabular} & No & \begin{tabular}[c]{@{}c@{}}Position;\\ Orientation\end{tabular}  & -  \\ \hline
    \begin{tabular}[c]{@{}c@{}}RumexWeeds\\ \cite{guldenring2023rumexweeds} \end{tabular} & Weed Detection &  \begin{tabular}[c]{@{}c@{}}Outdoors;\\ Over crop;\\ Grass\end{tabular} & Grass &  RGB & Yes & Yes & Yes & Position & Software  \\ \hline
    \begin{tabular}[c]{@{}c@{}} ARD-VO\\ \cite{crocetti2023ardvo} \end{tabular} & \begin{tabular}[c]{@{}c@{}}Localization;\\ Mapping\end{tabular} &  \begin{tabular}[c]{@{}c@{}}Outdoors;\\ Under crop;\\ Trees\end{tabular} & \begin{tabular}[c]{@{}c@{}}Localization;\\ Mapping\end{tabular} &  \begin{tabular}[c]{@{}c@{}}Stereo RGB;\\ Multispectral\end{tabular} & Yes & RTK & Yes & \begin{tabular}[c]{@{}c@{}}Position;\\ Orientation\end{tabular}  & Software  \\ \hline
    \begin{tabular}[c]{@{}c@{}}Bacchus Long-Term\\ \cite{polvara2023bacchuslongtermblt}\end{tabular} & \begin{tabular}[c]{@{}c@{}}Localization;\\ Mapping\end{tabular} &  \begin{tabular}[c]{@{}c@{}}Outdoors;\\ Under crop;\\ Trees\end{tabular} & Grapevine &  \begin{tabular}[c]{@{}c@{}}RGB;\\ Depth\end{tabular} & Yes & RTK & Yes & Position & Software  \\ \hline
    \begin{tabular}[c]{@{}c@{}}Terrasentia Under-Canopy\\ \cite{cuaran2024terrasentiaundercanopy}\end{tabular} & \begin{tabular}[c]{@{}c@{}}Localization;\\ Mapping\end{tabular} &  \begin{tabular}[c]{@{}c@{}}Outdoors;\\ Under crop;\\ Crop rows\end{tabular} & \begin{tabular}[c]{@{}c@{}}Corn;\\ Soybean\end{tabular} &  Stereo RGB & Yes & Single-Point & Yes & Position & Software  \\ \hline
    \begin{tabular}[c]{@{}c@{}}MAgro\\ \cite{marzoatanco2024magrodataset} \end{tabular} & \begin{tabular}[c]{@{}c@{}}Localization;\\ Mapping\end{tabular} &  \begin{tabular}[c]{@{}c@{}}Outdoors;\\ Under crop;\\ Trees\end{tabular} & \begin{tabular}[c]{@{}c@{}}Apple;\\ Pear\end{tabular} &  Stereo RGB & Yes & RTK & Yes & Position & Software  \\ \hline
    \begin{tabular}[c]{@{}c@{}}BotanicGarden\\ \cite{liu2024BotanicGarden} \end{tabular} & \begin{tabular}[c]{@{}c@{}}Localization;\\ Mapping;\\ Segmentation\end{tabular} &  \begin{tabular}[c]{@{}c@{}}Outdoors;\\ Under crop;\\ Trees\end{tabular} & Unspecified &  Stereo RGB & Yes & No & Yes & \begin{tabular}[c]{@{}c@{}}Position;\\Orientation\end{tabular} & Hardware  \\ \hline
    \begin{tabular}[c]{@{}c@{}}The GREENBOT Dataset\\ \cite{canadasaranega2024greenbotdataset}\end{tabular} & \begin{tabular}[c]{@{}c@{}}Localization;\\ Mapping\end{tabular} &  \begin{tabular}[c]{@{}c@{}}Indoors;\\ Under crop;\\ Crop rows\end{tabular} & Tomato &  Stereo RGB & Yes & No & No & - & - \\ \hline
    \textbf{Ours} & \begin{tabular}[c]{@{}c@{}} Localization;\\ Mapping \end{tabular} & \begin{tabular}[c]{@{}c@{}}Outdoors;\\ Over crop;\\ Crop rows\end{tabular}  & Soybean &  \begin{tabular}[c]{@{}c@{}}Stereo IR;\\ RGB; Depth\end{tabular}                     & Yes & \begin{tabular}[c]{@{}c@{}}Single-Point;\\ RTK; PPK\end{tabular} & Yes      & \begin{tabular}[c]{@{}c@{}}Position;\\ Orientation\end{tabular}  & \begin{tabular}[c]{@{}c@{}}Hardware +\\ Software\end{tabular} \\ \hline
\end{tabular}%
}
\caption{Publicly available datasets that target robot navigation and localization in agricultural environments (non-synthetic). A dash (-) means that information could not be obtained about that particular characteristic.}
\label{tab:dataset_survey}
\end{table*}

\section{Platform} \label{sec:platform}
This section describes the physical setup of sensors and actuators in the robotic platform as well as details of the software used to capture the data and the calibration methodology applied.
\subsection{The Robot}
The robot consists of a mobile platform with four independently driven wheels as depicted in Figure~\ref{fig:weed_removing_robot}.
It has been designed with autonomy in agricultural settings in mind, and as such it's powered by batteries charged by a series of solar panels in the top of the vehicle, and the distance and height of the wheels is consistent with the row distance and height of certain crops such as soy.

The aim of the robot as a whole is to automate the weeding of large crop fields. Particularly we aim to automate the robot's traversal through a cultivated field whilst classifying the images and applying removal techniques for weeds or enriching the crops with fertilizer (things currently in testing). The robot motion is controlled by four independent brushless motors (one per wheel). For the front wheels direction, a stepper motor has been built with the appropriate reduction and encoder. For this dataset the motion of the platform was performed 
by a human operator on-site with a radio-frequency controller.

\subsection{Data Collection System Overview}
\begin{figure}[!htbp]
    \centering
    \includegraphics[width=\linewidth]{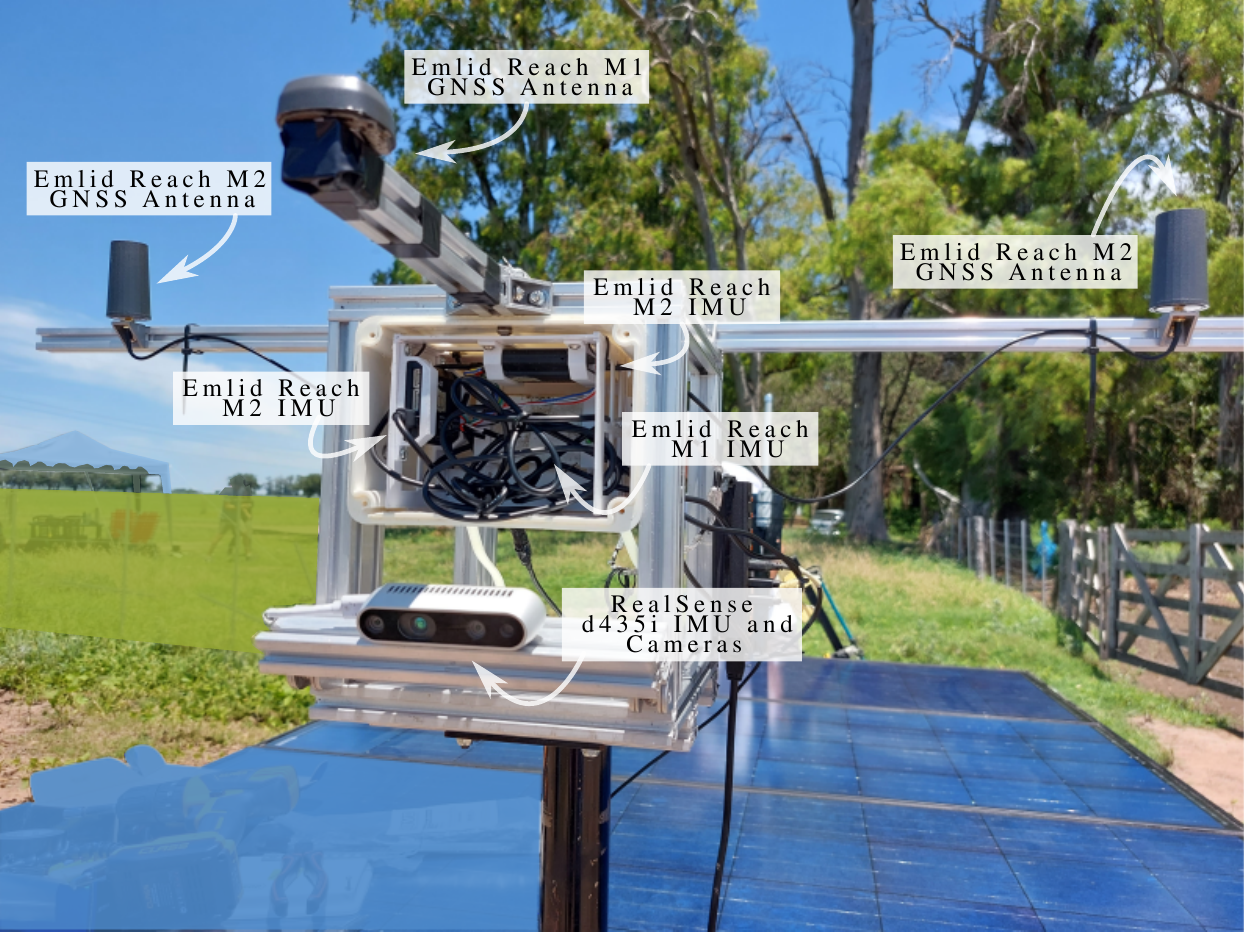}
    \caption{Placement of the different sensors in the sensor box.}
    \label{fig:sensor_box_setup}
\end{figure}

Figure~\ref{fig:sensor_box_setup} shows the sensor's placement in the box mounted at the front of the robot.
The exact positioning, orientation and frames of reference are available in the files of the dataset or the repository.
The setup consists of the sensors listed in Table~\ref{tab:sensor_list}, which provide both exteroceptive and proprioceptive data on the platform's environment and state.
\begin{table*}[!htbp]
    \centering
    \begin{tabular}{cccc}
    \hline
    \textbf{Name} &
      \textbf{Sensor} &
      \textbf{\begin{tabular}[c]{@{}c@{}}Resolution /\\ Range\end{tabular}} &
      \textbf{Acquisition Rate} \\ \hline
    \multirow{6}{*}{Intel Realsense D435i} &
      Stereo IR Camera &
      \begin{tabular}[c]{@{}c@{}}1280px × 720px\\ 87° × 58°\end{tabular} &
      \SI{15}{\hertz} \\ \cline{2-4} 
     &
      Color Camera &
      \begin{tabular}[c]{@{}c@{}}1280px x 720px\\ 69° × 42°\end{tabular} &
      \SI{15}{\hertz} \\ \cline{2-4} 
                                        & 6-DoF IMU      & \begin{tabular}[c]{@{}c@{}}± \SI{4}{\gram}\\ ± \SI{1000}{\deg\per\second}\end{tabular} & \SI{200}{\hertz} \\ \hline
    \multirow{4}{*}{Emlid Reach M1}     & GNSS     & ± \SI{2.5}{\meter}                                                     & \SI{5}{\hertz}   \\ \cline{2-4} 
                                        & RTK-GNSS & ± \SI{0.04}{\meter}                                                     & \SI{5}{\hertz}   \\ \cline{2-4} 
                                        & 9-DoF IMU      & \begin{tabular}[c]{@{}c@{}}± \SI{8}{\gram}\\ ± \SI{1000}{\deg\per\second}\end{tabular}    & \SI{200}{\hertz} \\ \hline
    \multirow{4}{*}{Emlid Reach M2 (1)} & GNSS     & ± \SI{2.5}{\meter}                                                      & \SI{5}{\hertz}   \\ \cline{2-4} 
                                        & RTK-GNSS & ± \SI{0.04}{\meter}                                                     & \SI{5}{\hertz}   \\ \cline{2-4} 
                                        & 9-DoF IMU      & \begin{tabular}[c]{@{}c@{}}± \SI{8}{\gram}\\ ± \SI{1000}{\deg\per\second}\end{tabular}    & \SI{200}{\hertz} \\ \hline
    \multirow{4}{*}{Emlid Reach M2 (2)} & GNSS     & ± \SI{2.5}{\meter}                                                      & \SI{5}{\hertz}   \\ \cline{2-4} 
                                        & RTK-GNSS & ± \SI{0.04}{\meter}                                                     & \SI{5}{\hertz}   \\ \cline{2-4} 
                                        & 9-DoF IMU      & \begin{tabular}[c]{@{}c@{}}± \SI{8}{\gram}\\ ± \SI{1000}{\deg\per\second}\end{tabular}    & \SI{200}{\hertz} \\ \hline
    \begin{tabular}[c]{@{}c@{}}E-Bike Wheel\\ Odometer\end{tabular} &
      Hall-Effect Odometry &
      ± \SI{7.5}{\degree} &
      \SI{10}{\hertz} \\ \hline
    \begin{tabular}[c]{@{}c@{}}OMRON \\ E6CP-A\end{tabular} &
    \begin{tabular}[c]{@{}c@{}}8-bit Absolute Encoder\end{tabular} &
    \begin{tabular}[c]{@{}c@{}}± \SI{1.4}{\degree}\\ \SI{92}{\degree}\end{tabular} &
      \SI{10}{\hertz} \\ \hline
    \end{tabular}
    \caption{List of sensors mounted on the platform and their characteristics.}
    \label{tab:sensor_list}
\end{table*}

To capture the data, the sensors are connected to a NVIDIA Jetson Orin Nano \SI{8}{\giga\byte} Development Kit computer with the following connections:
\begin{itemize}
    \item the Intel Realsense D435i is connected to a USB-A 3.0 port,
    \item the Reach GNSS modules (all three) are connected to a powered USB 3.0 hub, that in turn is connected to a USB-A 3.0 port,
    \item a single Emlid Reach GNSS module from the above (an M2 module) is connected also to a set of GPIO pins to provide a PPS (\emph{Pulse-per-second}) timing capability.
\end{itemize}
A diagram of the connections can be seen in Figure~\ref{fig:synchronization_scheme}.

We also make use of a Emlid Reach M1 GNSS module as a static base for our RTK solution, placing it close to the field we were working on.
The base sends RTK corrections, transmitted to the robot's GNSS modules through a long range Wi-Fi setup, to improve the accuracy of our GNSS positioning.

\subsection{Software Environment}
The data capture system is powered by ROS, where a set of ROS Nodes transform the raw sensor data into ROS Messages and then collects it into a ROS Bag.

We used the official package\footnote{\url{https://github.com/IntelRealSense/realsense-ros}}  provided by the manufacturer to convert the Intel Realsense D435i images and IMU measurements to ROS. For the Reach GNSS sensors, a slightly modified version of the Reach RS Driver node\footnote{\url{https://github.com/enwaytech/reach_rs_ros_driver}} does the   transformation from NMEA sentences into the standard ROS NavSatFix Message. There is no official software to obtain IMU data from the Reach modules, so we run RTIMULIB2\footnote{\url{https://github.com/HongshiTan/RTIMULib2}} on the modules to retrieve raw data (accelerometer, gyroscope and magnetometer) and transmit it to the host, which transforms that information into the corresponding ROS messages.

As for the odometry we record Hall effect sensor pulses for each of the four wheels and the steering system. The motors' microcontroller transmits to the host the instant velocity of the wheels in \si{\rpm}, the angle sensed by the encoder and the movement's direction (forward/backward).
To measure the angle of rotation of the wheels, a variable of interest for vehicle control, an 8-bit absolute encoder (OMRON E6CP-A) was used, mounted on the same stepper motor. With that information the wheel odometry can be obtained by the Ackermann model \cite{mueller2019modern}. The wheelbase of our robot is $\SI{1.65}{\meter}$, the steering angle $\in \left[-0.72,0.58\right] \si{\rad}$ and the wheel diameter $\SI{0.57}{\meter}$. The dataset includes the post-processed odometry, robot linear and angular velocities, and the raw  data.
The embedded system controls the wheel motors and their steering.
\subsection{Synchronization}

\begin{figure}[!htbp]
    \centering
    \includegraphics[width=\linewidth]{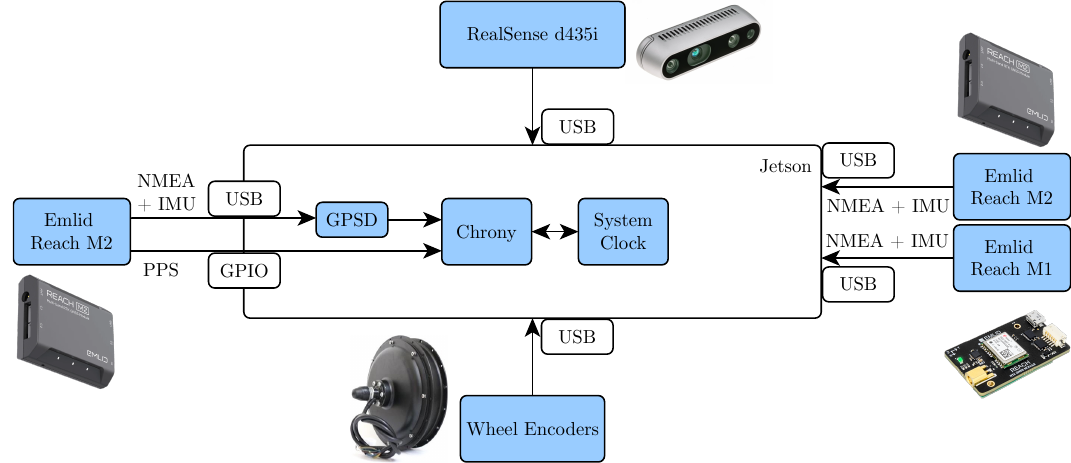}
    \caption{Synchronization scheme, where a single Emlid Reach M2 sensor is used with its PPS signal to synchronize the Jetson's OS clock with the GNSS UTC time. This time reference is then propagated to the RealSense D435i allowing referencing acquired sensor data with respect to a unique time source.}
    \label{fig:synchronization_scheme}
\end{figure}

In isolated environments, such as agricultural fields, mobile network coverage is often unavailable. In such scenarios, it is required to time synchronize the integrated computing system with the GNSS modules in order to reference acquired sensor data with respect to a unique time source, as GNSS modules are able to converge with the satellite's Coordinated Universal Time (UTC). The GNSS UTC time information (NMEA message) provided by one of the Emlid Reach GNSS M2 module is combined with its PPS (Pulse-per-Second) signal to set the clock of the Linux system \footnote{Ubuntu 18.04.6 LTS (GNU/Linux 4.9.140-tegra aarch64)} running in the NVIDIA Jetson module. The Linux OS uses the ``Chrony''~\footnote{\url{https://chrony-project.org/}} Network Time Protocol (NTP) implementation for this purpose, which makes the system time to gradually converge with the GNSS time by slowing down or speeding up the clock as required taking into account the GNSS time, the PPS signal and transmission latency that might exist. Figure \ref{fig:synchronization_scheme} depicts the described data synchronization scheme.
This configuration allows precise clock synchronization that is forwarded to the entire software environment aforementioned. In this way, ROS nodes directly timestamp and record messages in UTC reference. It is worth mentioning that inertial information provided by IMUs inside the Emlid Reach modules are directly generated in UTC time reference. Furthermore, the Intel RealSense D435i sensor unit is time synchronized with the Jetson system clock making use of its ``Global Time Domain'' driver feature which adjusts the D435i sensor clock considering the USB transmission latency. This allows the D435i sensor to directly generate data frames timestamped in the UTC time reference. The wheel odometry sensors are the only ones whose data is only synchronized by software, this is, we timestamp the hall-effect readings when they arrive to the Jetson module and are subsequently converted into ROS messages.

This schema improves over the purely software-based synchronization present in the previous Rosario Dataset, given that it allows all sensor readings to easily be synchronized with one another and with the readings of the GNSS measurements.
This also makes it straightforward to compare results obtained with a SLAM system against the ground-truth data provided by the GNSS measurements, without the need to manually adjust the timing by matching measurements of different time sources.

\subsection{Calibration}

\subsubsection{IMU Intrinsic Calibration:}
\begin{figure}[!htbp]
    \centering
    \includegraphics[width=\linewidth]{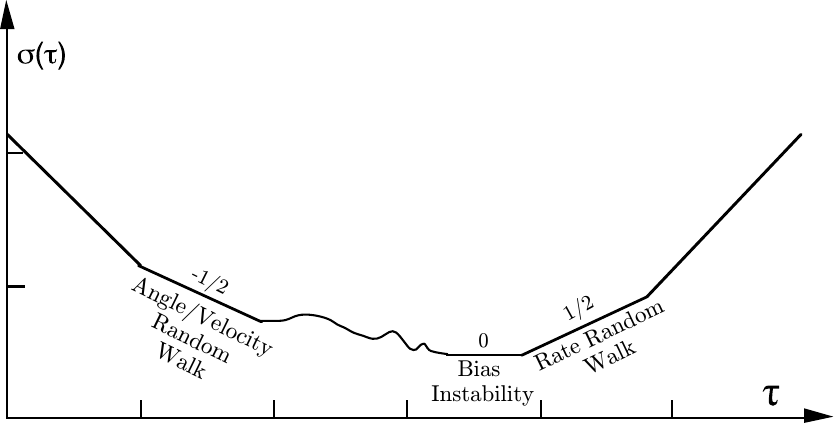}
    \caption{Allan deviation $\sigma$ in function of $\tau$. This plot allows to identify different types of noise.}
    \label{fig:adev_plot}
\end{figure}

IMUs suffer from various stochastic errors, such as bias instability and random walk, which can degrade the accuracy of navigation and mapping applications. Intrinsic calibration of IMUs is required to determine internal sensor parameters like scale factors and axis misalignments, crucial for accurate motion tracking, especially in applications requiring high precision sensor performance. Allan variance is a statistical tool widely used to assess the stability of sensor measurements over time, helping to identify noise characteristics and systematic errors. The Allan variance is computed by segmenting the IMU data into $N$ clusters of a given averaging time $\tau$. Then, the average values of the signal over each segment is computed. The Allan Variance is then given by:
\begin{equation}
\sigma^{2}\left(\tau\right)=\frac{1}{2\left(n-1\right)}\sum_{i=1}^{N-1}\left(\bar{x}_{i+1}-\bar{x}_{i}\right)^{2},
\end{equation}
where $\bar{x}_{i}$ are the averaged values over segments \cite{woodman2007inertial}. We focus on its square root, the Allan deviation $\sigma$. To identify different noise types, we plot $\sigma$ as a function of $\tau$ on a log-log scale, where distinct noise characteristics appear as specific slope patterns \cite{ieee1998gyro}. Figure~\ref{fig:adev_plot} presents an example of this type of plot. We aim to extract the following parameters: gyroscope white noise (also known as gyroscope noise density or angular random walk), gyroscope random walk, accelerometer white noise (also known as accelerometer noise density or velocity random walk), and accelerometer random walk. White noise parameters are obtained by fitting a line to the region of the curve where the slope is $-\frac{1}{2}$ and reading its value at $\tau=1$. Conversely, gyroscope and accelerometer random walk parameters are determined by fitting a line in the region where the slope is $\frac{1}{2}$ and reading its value at $\tau=3$.

To accomplish this, we recorded data for \SI{3}{\hour} from the three IMUs of the Emlid Reach modules as well as from the RealSense IMU. Next, we compute the Allan deviation for both the gyroscope and accelerometer using the Allan Variance ROS Toolbox~(\cite{allanvarianceros}) and extract the four key parameters from the resulting plots. Table~\ref{tab:imu_calibration_parameters} shows the results. These parameters are required to calibrate camera and IMU jointly using Kalibr.
\subsubsection{Camera-IMU Calibration:}
Once we obtain the aforementioned parameters from the Allan Variance method we can jointly calibrate the stereo camera and IMUs. For this step we use Kalibr~(\cite{redher2016kalibr}). After collecting enough visual-inertial measurements to perform the calibration, we run Kalibr on this data and supply the IMU intrinsic parameters as input. As a result of the calibration, we obtained the camera intrinsic parameters and the extrinsic transformations between the stereo camera and the IMUs.

\subsubsection{Odometry Calibration:}

\begin{figure}[!htbp]
    \centering
    \includegraphics[width=\linewidth]{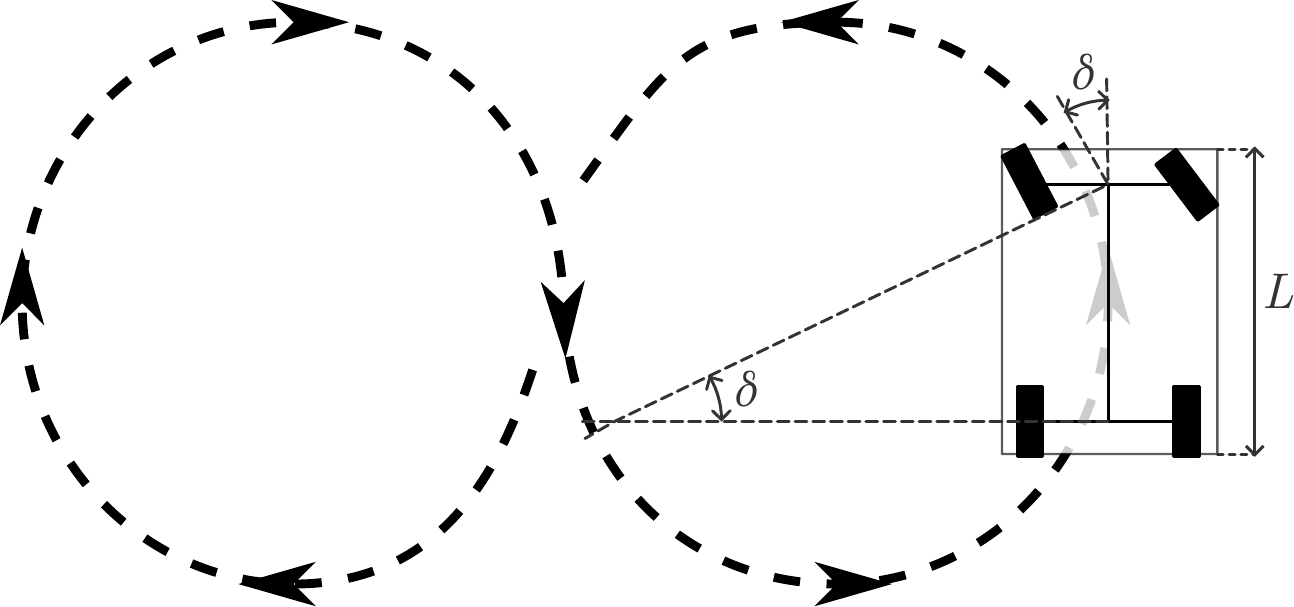}
    \caption{Diagram of the trajectory of the calibration routine and parameters estimated by the odometry calibration.}
    \label{fig:calibration_trajectory}
\end{figure}

The Ackermann model parameters were manually determined for wheel diameter and wheelbase ($L$), and semi-automatically for steering angles ($\delta$).
The procedure to obtain the steering angles consists of first determining the diameter of the ground-truth curves corresponding to the robot's maximum steering angles (hard right and hard left) and then using a physics-based simulation of the robot, we perform a parameter optimization by iteratively adjusting the angle inputs to minimize the cost function defined as the difference between the simulated and ground-truth trajectories.
This is depicted in Figure~\ref{fig:calibration_trajectory}.

\subsubsection{Extrinsic Calibration of Remaining Sensors:}
\begin{figure*}[!htbp]
    \centering
    \includegraphics[width=\linewidth]{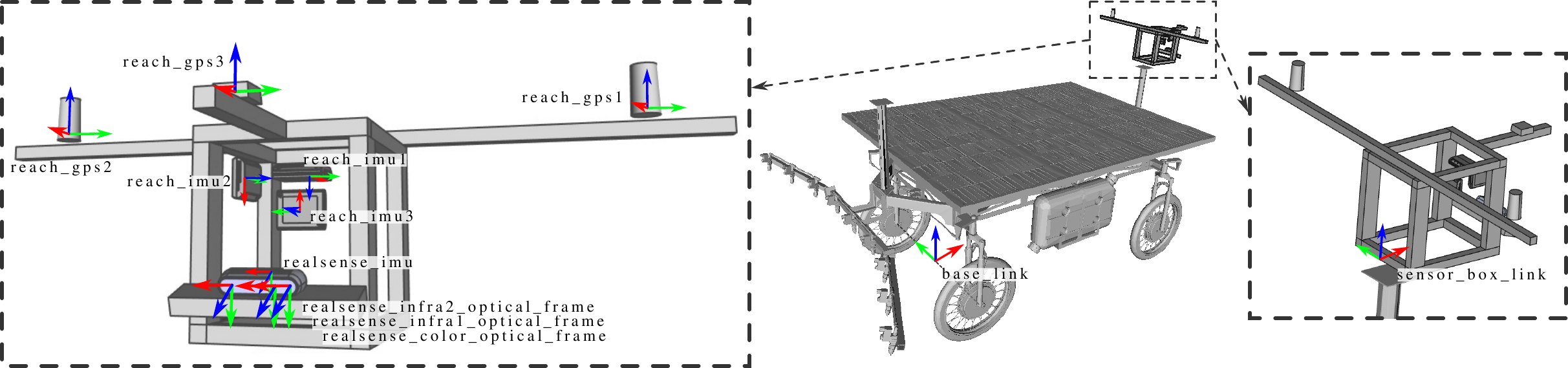}
    \caption{Transformations for the robot are shown overlaid onto a 3D representation of the robot, with two zoomed-in plots for the sensor box suite to better showcase its sensor and link frames. The RealSense optical frame names are stacked from left to right.}
    \label{fig:robot_tf}
\end{figure*}

Extrinsic measurements for the remaining sensors were taken manually.
Figure~\ref{fig:robot_tf} shows a 3D representation of the different sensors frames of reference.
The robot's \verb|base_link| was set at the mid-point of the rear wheel axes, and the \verb|sensor_box_link| was set at the contact point for the sensor suite box and the chassis, at the mid point of the rear bottom-most point of the box. 

\section{Data Collection} \label{sec:dataset}
This section provides an overview of the dataset collected by the robot in the field, along with calibration data for the sensors.
We analyze the different recorded sequences, presenting the ground-truth, statistics and environment conditions.

\begin{figure}[!htbp]
    \centering
    \subfloat[First Day (2023-12-22)]{\includegraphics[width=.49\linewidth]{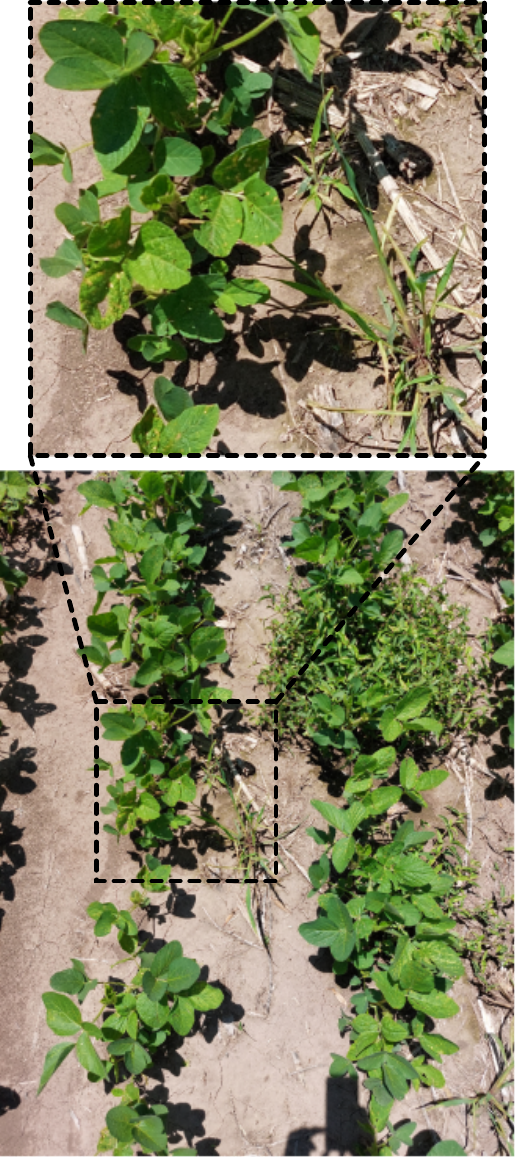}}
    \hfill
    \subfloat[Second Day (2023-12-26)]{\includegraphics[width=.49\linewidth]{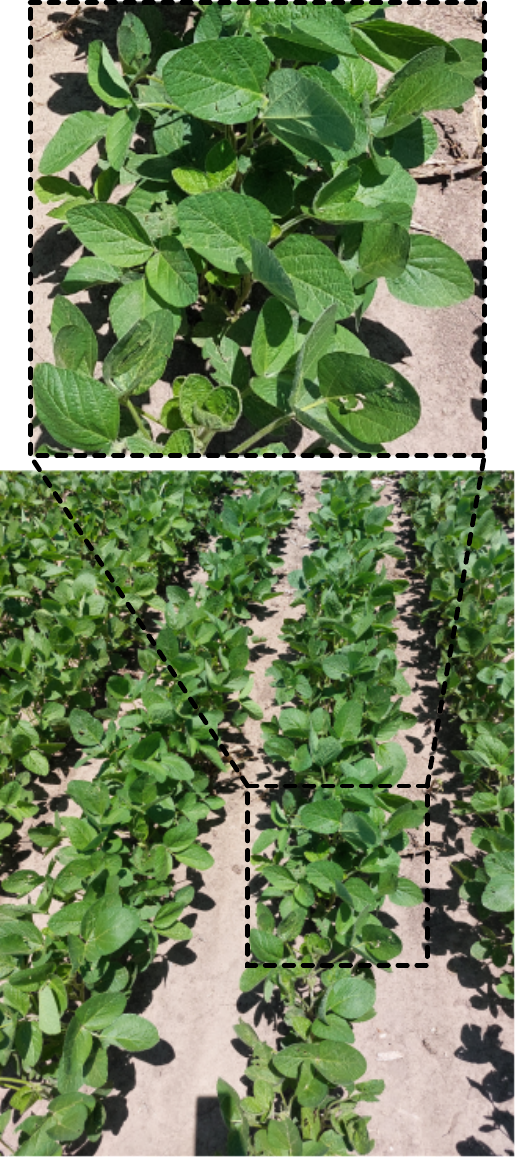}}
    \caption{Sample images of the soybean crops from the two days where the recordings took place. A zoom-in shows the plants in more detail, as well as some of the weeds present in the field. The images were taken with a handheld camera.}
    \label{fig:soybean_field}
\end{figure}

We recorded six separate sequences of the robot traversing a soybean plantation.
The first three sequences (sequences \#1 to \#3) took place on the December 22nd, 2023, in a field referred to as Field\#1.
The remaining three sequences (sequences \#4 to \#6) were conducted on the December 26th, 2023, in a different field, approximately \SI{830}{\meter} away from the first, which we will call Field\#2.
Both fields belong to the Faculty of Agronomy (National University of Rosario), located in Zavalla, Santa Fe, Argentina.
On the first day's field the rows of soybean plants were sowed with a spacing of \SI{0.52}{\meter}, whereas in the second day's field the rows were spaced by \SI{0.42}{\meter}, both with none or minimum tilling.
The plants were manually measured and had an average width of \SI{0.30}{\meter} and an average height of \SI{0.40}{\meter} on both dates.
The plants were in a vegetative growth stage of V6 to V8 in the first day, and V3 or V4 in the second day, where a vegetative stage of V$i$ means that the plant has $i$ trifoliate leaves completely expanded. The stage of V8 precedes the plant flowering stage.
Figure~\ref{fig:soybean_field} shows close-up images of the soybean plants and the disposition of the crop rows.

\begin{figure*}[!htbp]
    \centering
    \includegraphics[width=\linewidth]{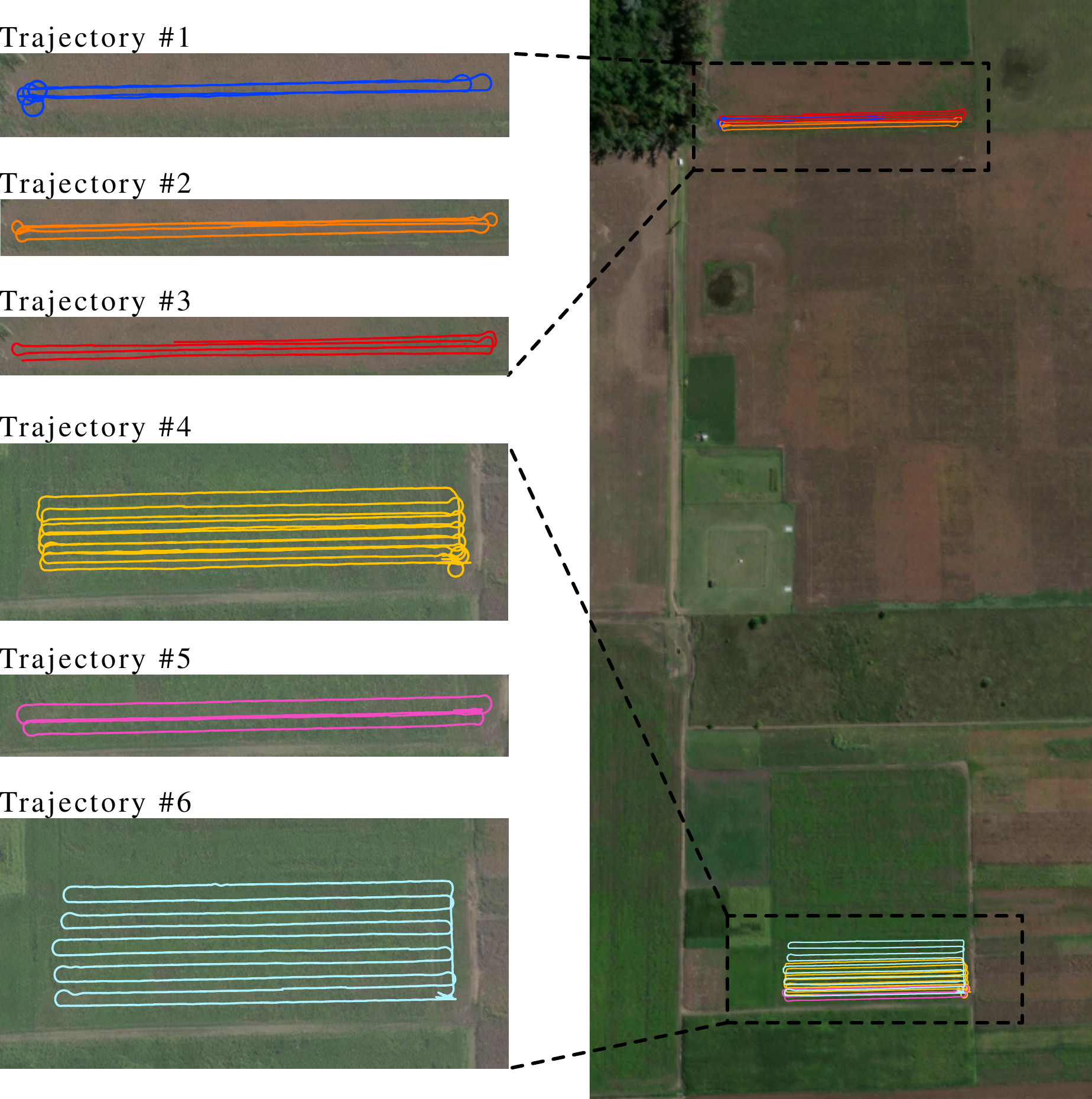}
    \caption{The six trajectories of the GNSS data from the dataset plotted onto a satellite map with zoomed out boxes singling out each trajectory individually.}
    \label{fig:map_and_trajectories}
\end{figure*}

Figure~\ref{fig:map_and_trajectories} shows the trajectories overlaid on a map, while Table~\ref{tab:sequences} summarizes the key aspects of the six sequences.
We recorded a total of \SI{2.23}{\hour} of trajectory, with a total length of over \SI{7.33}{\km}, more than four times the duration of the previous Rosario Dataset, and covering more than three times the distance.

At the beginning of sequences \#1 and \#4 (the first sequence for each distinct field) we performed a calibration routine. This involved maneuvering the robot forwards and backwards three times in a straight line and then making circular patterns (first turning fully clockwise and then turning fully counter-clockwise) while capturing images of two Kalibr AprilTag grids. Such initial calibration pattern guarantees to have data across all axes in the 2D plane allowing, for instance, wheel odometry and magnetometer sensor calibration and SLAM system initialization. 
The dataset includes metadata that marks the time when the calibration ends, allowing users to ignore this portion if they are not interested in the calibration data.

\begin{figure*}[!htbp]
    \centering
    \includegraphics[width=\linewidth]{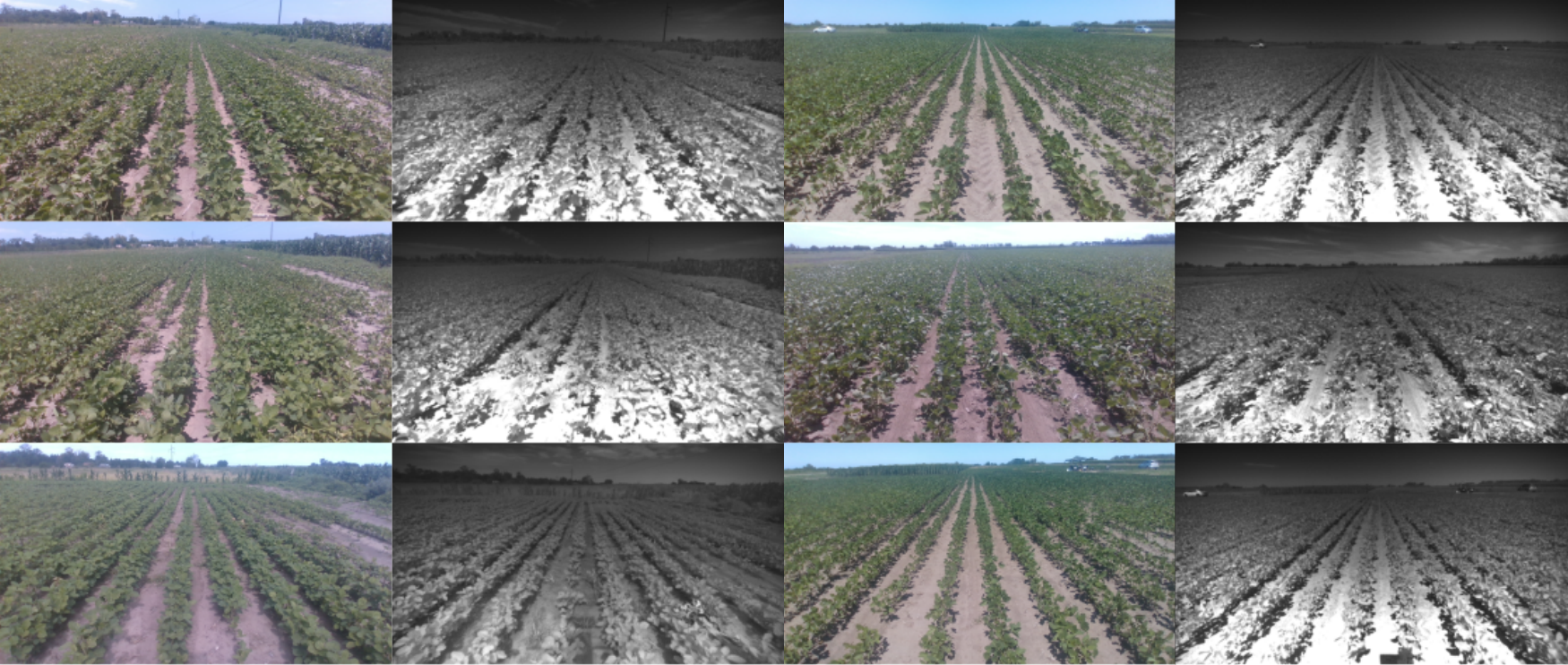}
    \caption{Sample images from the RealSense D435i color camera and stereo infrared camera (left camera only), one pair for each trajectory.}
    \label{fig:sample_images}
\end{figure*}

The recorded sequences were planned to contain overlapping trajectories, both in a given sequence and in between different sequences, to allow for testing of loop detection and loop closure systems in these environments.
We also made sure to record on sunny days, which provides good natural illumination, an important factor for key-point detection and a good system localization.
Other characteristics of the sequences are that they always move forwards, except when we perform the calibration routines, as a typical agriculture machine would do, and that the terrain is rough, which presents some naturally occurring shakiness in the camera feed and the sensor readings.
Figure~\ref{fig:sample_images} shows sample images from the stereo IR camera and the color camera.

The robot's movement was manually controlled by an operator with a radio-frequency controller as described in Section~\ref{sec:platform}.

\begin{table*}[!htbp]
    \centering
    \begin{tabular}{cccccc}
    \hline
    Sequence & Sequence ID & Duration (\unit{\second}) & Distance (\unit{\meter}) & Calibration Start (\unit{\second}) & Calibration End (\unit{\second}) \\ \hline
    \#1 & 2023-12-22-13-14-16 & 940  & 777  & 0  & 218 \\
    \#2 & 2023-12-22-14-29-43 & 1011 & 904  & -  & -   \\
    \#3 & 2023-12-22-16-31-08 & 943  & 950  & -  & -   \\
    \#4 & 2023-12-26-13-39-43 & 2506 & 2254 & 0  & 190 \\
    \#5 & 2023-12-26-15-10-15 & 796  & 703  & -  & -   \\
    \#6 & 2023-12-26-15-48-38 & 1862 & 1744 & -  & -   \\
    \multicolumn{1}{l}{} &  \multicolumn{1}{l}{} & \multicolumn{1}{l}{} & \multicolumn{1}{l}{} & \multicolumn{1}{l}{} & \multicolumn{1}{l}{}
    \end{tabular}
    \caption{Characteristics of the six recorded sequences of the dataset.}
    \label{tab:sequences}
\end{table*}

\subsection{Ground-Truth}
\label{sec:ground_truth}
In addition to the GNSS-RTK positional solution, we compute GNSS-PPK (Post-Processed Kinematic) solutions for each Emlid Reach GNSS module to achieve higher precision. All raw GNSS data was saved for both the base station and the modules on board the rover. RINEX files stored by the Emlid Reach GNSS modules where post-process with the RTKLIB\footnote{\url{https://github.com/rtklibexplorer/RTKLIB.git}}, a open-source library for processing GNSS data that provides tools for computing GNSS-PPK solutions. We performed this procedure for each Emlid Reach device onboard the robot using all available corrections from the base station module.

The 6-DoF inertial information provided by the Intel Realsense D435i IMU was later fused with the GNSS+PPK localization solutions using the MINS \cite{lee2023mins} multisensor fusion framework. Using the two GNSS-PPK solutions from the Emlid Reach M2 modules, MINS is able to resolve a globally referenced orientation in a East-North-Up (ENU) global frame of the Intel Realsense D435i IMU coordinate frame system for every inertial reading. Measurements from the Emlid Reach M1 module were discarded from the computation due to its difficulty in acquiring a fix and its inherent limitations as a single-band receiver. During real-time operation, the M1 experienced poor SNR (signal-to-noise ratio) and struggled to maintain a fix for a few minutes, resulting in measurements that were unreliable for fusion with the data from the other sensors. Ground-truth localization information is finally generated applying the extrinsic transformation between the IMU frame and the robot body frame (base\_link frame).

The estimated standard deviation of the PPK solutions, as reported in the RTKLIB output during post-processing, is approximately 5 mm in the horizontal plane and 1 cm in altitude. These values are derived from the covariance reported by the internal EKF of RTKLIB and provide a reasonable indication of internal confidence in the solution. Furthermore, both PPK solutions used as input to MINS maintained a Fix status \SI{100}{\percent} of the time, ensuring maximum confidence in their positional accuracy. When comparing the MINS output with one of the original PPK trajectories, the mean positional difference varies across sequences, typically ranging between \SI{4}{\cm} and \SI{8}{\cm}. Notably, MINS provides a reliable estimate of orientation, which is not available from PPK alone. We are releasing both the original PPK solutions and the MINS-derived trajectories as part of the dataset, this ensures that users can choose the most suitable reference for their needs and the possibility of performing their own sensor fusion and orientation estimation if desired.

\section{Experimental Evaluation} \label{sec:results}
\begin{figure}[!hbtp]
    \centering
    \includegraphics[width=\linewidth]{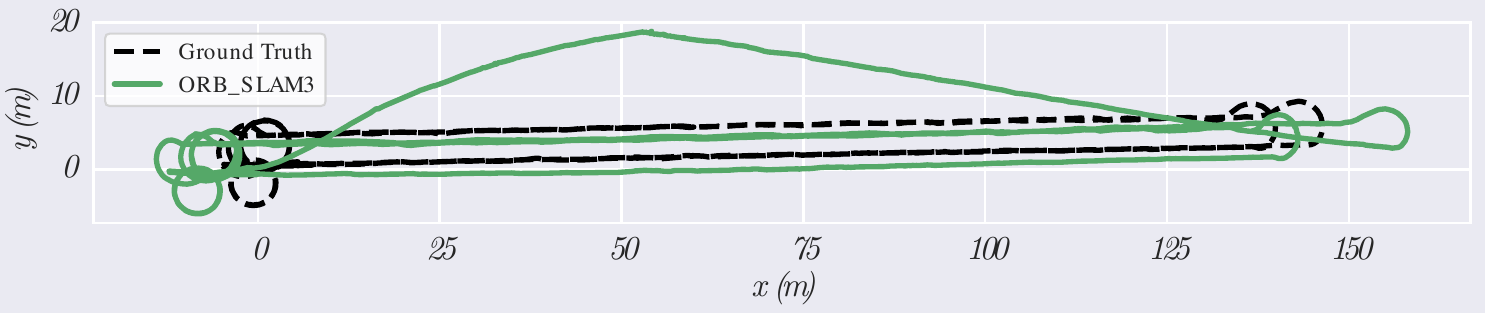}
    \caption{Trajectory obtained by running ORB-SLAM3 on sequence \#1 with loop closure active. The system incorrectly closes 7 loops and the estimated trajectory gets irreversibly corrupted from them.}
    \label{fig:trajectory1_loop_closure}
\end{figure}

To validate the proposed dataset, we evaluate and analyze state-of-the-art multi-modal SLAM systems. In particular, we assess the stereo-inertial configuration of ORB-SLAM3~(\cite{campos2021orbslam3}), the GNSS-stereo-inertial configuration of ORB-SLAM3+GNSS~(\cite{cremona2023gnss}) and the stereo-inertial implementation of OpenVINS~(\cite{geneva2020openvins}).
For ORB-SLAM3 we set the same parameters as in \cite{cremona2022evaluation}, as they also evaluate the method on the previously released agricultural dataset ``The Rosario Dataset''.
Also for ORB-SLAM3+GNSS we use the default parameters provided in the implementation released by \cite{cremona2023gnss}, which was also tested on the former dataset.
In the case of OpenVINS we set similar parameters to the other methods: 1200 features extracted per image, with a state vector of 50 points and KLT tracking (as recommended by the authors).
We use the RealSense D435i IR stereo camera and its IMU, and a single dual-band GNSS (an Emlid Reach M2) to run the systems. The results show that, initially, all systems estimate the robot trajectory properly, but as the robot navigates for a while, pose estimation drifts or even loses localization.
Although each system has a loop detection module, we turn off the module to avoid false positive loops resulting from the perceptual aliasing nature of the environment, as reviewed in \cite{soncini2024addressing}.
To justify this decision we show an example trajectory obtained from running ORB-SLAM3 with loop closure enabled in Figure~\ref{fig:trajectory1_loop_closure}.
In this trajectory the system closed 7 loops, with negligible translation errors and maximum rotation error of \SI{170}{\degree} (when comparing to ground-truth pose differences between the selected keyframes), irreversibly corrupting the estimated trajectory.

\begin{figure*}[!hbtp]
    \centering
    \includegraphics[width=\linewidth]{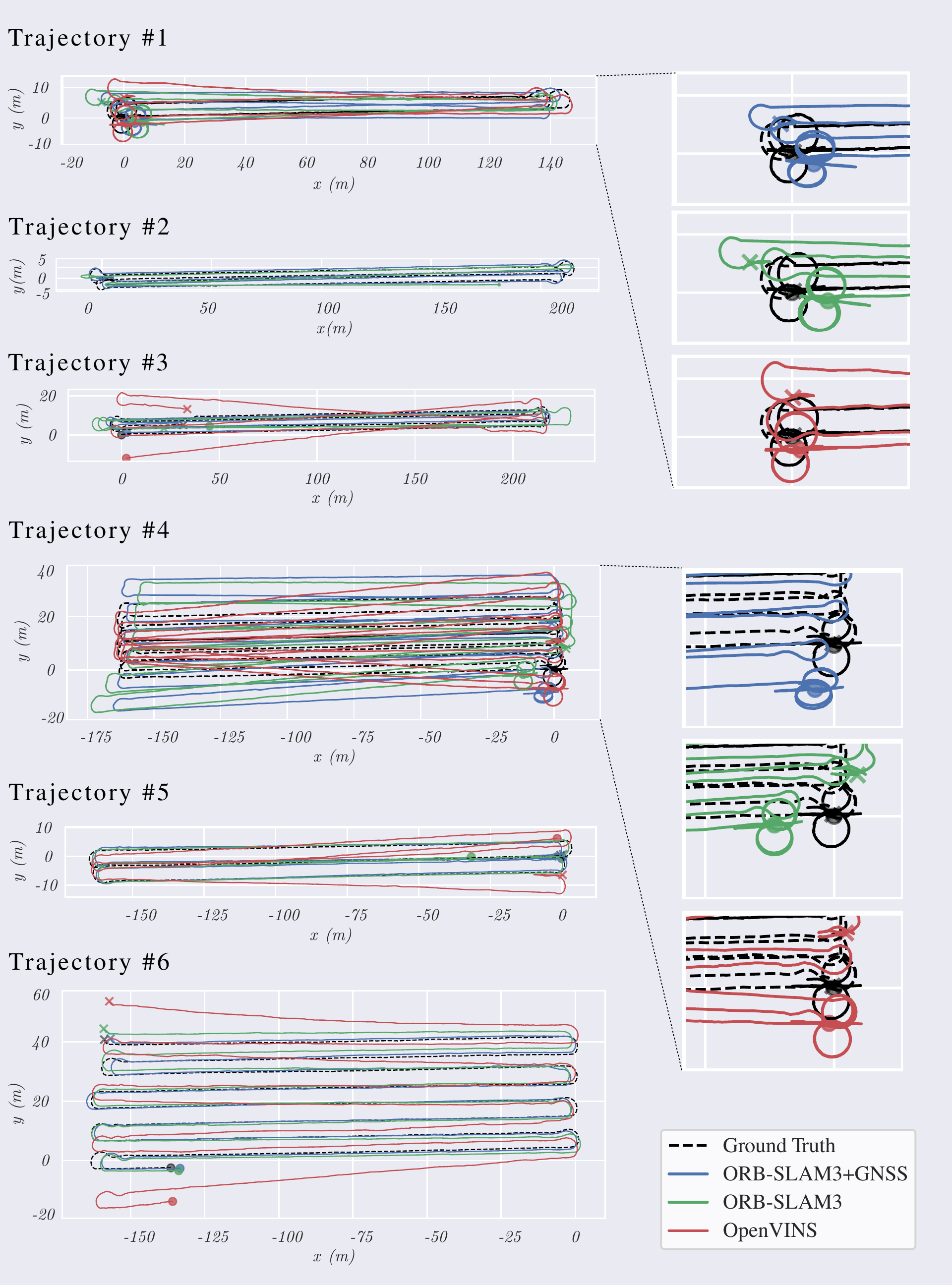}
    \caption{Trajectories obtained by running diverse SLAM systems on the six sequences of the dataset. Some results are shown partially or not shown at all for lack of good pose tracking by such methods. Zoomed in portions of trajectories \#1 and \#4 are shown around the initial and final positions of each method as the robot was purposely made to start and finish on the same spot.}
    \label{fig:trajectories_and_zooms}
\end{figure*}

Figure~\ref{fig:trajectories_and_zooms} shows the trajectories estimated by each system as the best alignment possible with the ground-truth trajectory via Umeyama alignment \cite{umeyama1991pointpatterns}.
OpenVINS did not run on sequence \#2 as it failed to initialize the IMU for tracking.
We omit the first few seconds for ORB-SLAM3 in sequences \#2, \#3, \#5 and \#6, as it takes some time for it to initialize correctly. In particular, sequence \#2 takes the longest, at around \SI{230}{\second} to converge.
ORB-SLAM3 also loses track in sequence \#6 with about \SI{300}{\second} left to finish.
It can be noted in the trajectories how the different methods drift over time, with the one showing some of the best trajectories being ORB-SLAM3+GNSS, as expected given it's the only one that fuses information from a GNSS sensor.
The figures also show the start and end poses of the robot for each trajectory as a similarly colored circle and cross respectively, where some zoomed-in clippings for trajectories \#1 and \#4 are shown, as the path for the robot was planned in such a way that the robot started and ended in the exact same place by adding stakes with red markers on the ground where the four wheels should end up.
We can see by the results that in these two trajectories none of the methods were able to perfectly start and end at the same time due to the accumulated drift.
It's our expectation that these issues could be mitigated by incorporating a loop closure system that works in agricultural fields, which is one of the motivations for releasing this dataset.

We evaluate the accuracy of the estimated trajectories of each method by calculating two standard metrics \cite{sturm2012benchmarkslam}: the absolute pose error (APE) (also called absolute trajectory error or ATE) which compares the absolute distances between the estimated and ground-truth trajectories, and the relative pose error (RPE) which measures the local accuracy of the trajectory over a fixed interval (which in our case was set to \SI{1}{\meter} displacement).
Since our sensors are time-synchronized it's easy to pair the poses of the estimated trajectories with our ground-truth, we still need to align both trajectories to a common frame of reference, which is done with Umeyama alignment \cite{umeyama1991pointpatterns}.

Table~\ref{tab:trajectories_ape} shows the APE results, where it can be seen that methods significantly drift on all sequences of the dataset, with particular issues when comparing the orientations of the trajectories.
A small amount of drift is expected for ORB-SLAM3 and OpenVINS given that these methods don't make use of any global information (e.g. GNSS, compass, or loop-closure systems), so as to make a fairer comparison we showcase in Table~\ref{tab:trajectories_rpe} the RPE metrics calculated over the SLAM trajectories and ground-truth, which only makes local comparisons over the estimated motion of the vehicle.
These results shows that agricultural environments, despite its importance, are still challenging for state-of-the-art SLAM systems.
\begin{table*}[!hbtp]
\centering
\resizebox{\textwidth}{!}{%
\begin{tabular}{ccccccc}
\hline
\multirow{2}{*}{Method} & \multicolumn{6}{c}{Sequence}  \\ \cline{2-7} 
 & \#1 & \#2 & \#3 & \#4 & \#5 & \#6 \\ \hline
\multirow{2}{*}{ORB-SLAM3} & \SI{5.17}{\meter} (2.44) & \SI{4.27}{\meter} (1.88)* & \SI{7.06}{\meter} (3.49)* & \SI{7.91}{\meter} (3.72) & \SI{1}{\meter} (0.54)* & \SI{1.83}{\meter} (0.88)* \\
 & \SI{25.05}{\degree} (0.94) & \SI{73.71}{\degree} (0.36)* & \SI{2.68}{\degree} (0.91)* & \SI{26.72}{\degree} (1.94) & \SI{11}{\degree} (0.26)* & \SI{1.64}{\degree} (0.6)* \\ \hline
\multirow{2}{*}{ORB-SLAM3+GNSS} & \SI{4.31}{\meter} (1.97) & \SI{1.83}{\meter} (0.90) & \SI{7.16}{\meter} (1.22) & \SI{7.12}{\meter} (3.94) & \SI{1.87}{\meter} (0.7) & \SI{1.86}{\meter} (0.99) \\
 & \SI{40.67}{\degree} (1.19) & \SI{29.89}{\degree} (0.55) & \SI{34.96}{\degree} (0.19) & \SI{49.77}{\degree} (1.62) & \SI{9.01}{\degree} (0.96) & \SI{17.63}{\degree} (0.73) \\ \hline
\multirow{2}{*}{OpenVINS} & \SI{2.30}{\meter} (1.73) & - & \SI{4.47}{\meter} (3.04) & \SI{3.48}{\meter} (2.53) & \SI{2.37}{\meter} (1.7) & \SI{4.37}{\meter} (3.11) \\
 & \SI{4.91}{\degree} (0.83) & - & \SI{18.08}{\degree} (0.35) & \SI{9.5}{\degree} (0.66) & \SI{12.43}{\degree} (0.31) & \SI{8.8}{\degree} (0.34) \\ \hline
\end{tabular}%
}
\caption{Mean and standard deviation (in parentheses) of the absolute pose error (APE) of the position and orientation of the trajectories given by each method, when aligned to the ground-truth trajectory with Umeyama's method. Values preceded by an asterisk (*) were the result of truncated trajectories as described in the text.}
\label{tab:trajectories_ape}
\end{table*}

\begin{table*}[!hbtp]
\centering
\resizebox{\textwidth}{!}{%
\begin{tabular}{ccccccc}
\hline
\multirow{2}{*}{Method} & \multicolumn{6}{c}{Sequence} \\ \cline{2-7} 
 & \#1 & \#2 & \#3 & \#4 & \#5 & \#6 \\ \hline
\multirow{2}{*}{ORB-SLAM3} & \SI{0.04}{\meter} (0.03) & \SI{0.09}{\meter} (0.04)* & \SI{0.06}{\meter} (0.05)* & \SI{0.04}{\meter} (0.03) & \SI{0.04}{\meter} (0.02)* & \SI{0.04}{\meter} (0.03)* \\
 & \SI{0.20}{\degree} (1.19) & \SI{0.16}{\degree} (0.16)* & \SI{0.14}{\degree} (0.12)* & \SI{0.17}{\degree} (1.32) & \SI{0.12}{\degree} (0.12)* & \SI{0.15}{\degree} (0.15)* \\ \hline
\multirow{2}{*}{ORB-SLAM3+GNSS} & \SI{0.04}{\meter} (0.03) & \SI{0.08}{\meter} (0.05) & \SI{0.04}{\meter} (0.06) & \SI{0.03}{\meter} (0.02) & \SI{0.04}{\meter} (0.04) & \SI{0.04}{\meter} (0.04) \\
 & \SI{0.19}{\degree} (1.45) & \SI{0.16}{\degree} (1.2) & \SI{0.16}{\degree} (1.03) & \SI{0.18}{\degree} (1.14) & \SI{0.18}{\degree} (2.24) & \SI{0.16}{\degree} (0.15) \\ \hline
\multirow{2}{*}{OpenVINS} & \SI{0.04}{\meter} (0.03) & - & \SI{0.07}{\meter} (0.08) & \SI{0.05}{\meter} (0.04) & \SI{0.05}{\meter} (0.03) & \SI{0.05}{\meter} (0.04) \\
 & \SI{0.16}{\degree} (0.15) & - & \SI{0.16}{\degree} (0.11) & \SI{0.14}{\degree} (0.12) & \SI{0.13}{\degree} (0.13) & \SI{0.15}{\degree} (0.11) \\ \hline
\end{tabular}%
}
\caption{Mean and standard deviation (in parentheses) of the relative pose error (RPE) of the position and orientation of pairs of poses at a distance of \SI{1}{\meter} for the trajectories given by each method. Values preceded by an asterisk (*) were the result of truncated trajectories as described in the text.}
\label{tab:trajectories_rpe}
\end{table*}

\section{Summary and Future Work} \label{sec:conclusions}
We presented a calibrated, synchronized and tested dataset aimed at evaluating and progressing the scope of autonomy in agricultural environments.
It is our belief that this dataset is needed due to the scarcity of agricultural data for autonomous operation on crop fields, and will help advance many areas of robotics, control, and computer vision.
This is no empty claim, as we provide experimental evaluations showing that current localization and mapping methods are lackluster in these environments and need further improvement.
Our results show that most of the issues pointed out in \cite{cremona2022evaluation} are still present in the results produced by state-of-the-art SLAM systems and thus strengthens the importance of releasing the dataset and furthering the research on the subject matter.  

In the future we plan to keep working on enhancing the information available on the dataset by curating pose and orientation ground-truths, enhancing the calibration results, annotating the visual cues and evaluating existing methods to further the work of autonomy on agricultural settings. 
We also have plans to continue recording data on the field in coming years, making an attempt to capture different conditions of lighting, weather and crop growth stages, incorporating new sensor configurations, such as lidars and radars, and enhancing the measurement tools, such as more precise imu and gnss modules.


\begin{acks}
We specially thank Engr. Néstor Di Leo from the Land Management Chair of the Faculty of Agricultural Sciences of the National University of Rosario for giving us access to the agricultural field.
\end{acks}

\begin{funding}
This work was partially supported by Consejo Nacional de Investigaciones Científicas y Técnicas (Argentina) under grants PIBAA No.0042, AGENCIA I+D+i (PICT 2021‐570), and by Universidad Nacional de Rosario (PCCT‐UNR 80020220600072UR).
\end{funding}

\bibliographystyle{SageH}
\bibliography{bibliography}

\begin{sm}

\section{Appendix A: Calibration Parameters} \label{appendix:calibration_parameters}
For the RealSense D435i cameras we used a radial-tangential calibration model that calibrates intrinsic parameters and distortion coefficients for radial distortion and tangential distortion.
The parameters are listed in Table~\ref{tab:camera_calibration_parameters} for all cameras.

\begin{table*}[!hbtp]
\centering
\resizebox{\textwidth}{!}{%
\begin{tabular}{ccccccccc}
    \multirow{2}{*}{Camera} & Intrinsics &  &  &  & Distortion &  &  &  \\
     & $f_x$ & $f_y$ & $c_x$ & $c_y$ & $k_1$ & $k_2$ & $p_1$ & $p_2$ \\ \cline{2-9} 
    color & $890.4202$ & $895.5269$ & $633.5761$ & $375.3947$ & $0.0511$ & $-0.0881$ & $-0.0011$ & $-0.0004$ \\
    infra1 & $645.4064$ & $648.5756$ & $648.7339$ & $349.0376$ & $0.0008$ & $-0.0010$ & $-0.0005$ & $0.0004$ \\
    infra2 & $645.9158$ & $649.1492$ & $647.7810$ & $348.7846$ & $0.0005$ & $-0.0005$ & $-0.0005$ & $0$
\end{tabular}%
}
\caption{RealSense D435i camera intrinsic parameters obtained with calibration, given with a precision of 8 decimal points.}
\label{tab:camera_calibration_parameters}
\end{table*}

For the RealSense D435i IMU and the Emlid Reach IMUs we calibrated the Allan Variance, from which the values listed on Table~\ref{tab:imu_calibration_parameters} were obtained.

\begin{table*}[!hbtp]
\centering
\resizebox{.85\textwidth}{!}{%
\begin{tabular}{lllll}
    \multirow{2}{*}{Module} & Accelerometer &  & Gyroscope &  \\
     & Noise Density & Random Walk & Noise Density & Random Walk \\ \cline{2-5} 
    RealSense D435i & $0.00099438$ & $0.00004750$ & $0.00023249$ & $0.00000182$ \\
    Emlid Reach M1 & $0.00236037$ & $0.00016363$ & $0.00015210$ & $0.00001414$ \\
    Emlid Reach M2 (1) & $0.00335015$ & $0.00011784$ & $0.00016841$ & $0.00002311$ \\
    Emlid Reach M2 (2) & $0.00269792$ & $0.00026127$ & $0.00019469$ & $0.00001151$
\end{tabular}%
}
\caption{Noise values obtained from the calibration for the different IMU units, given with a precision of 8 decimal points.}
\label{tab:imu_calibration_parameters}
\end{table*}

\section{Appendix B: Transformations} \label{appendix:transformations}
Table~\ref{tab:transformations} lists the rigid transformations between the different coordinate frames involved in the system.
Transformations are given by a combination of a translation vector $(tx, ty, tz)$ and a rotation quaternion $(qx,qy,qz,qw)$ (scalar-last).

The Git repository contains a convenient launch file that allows the \verb|tf| module to automatically publish all these transformations for use with ROS.

\begin{table*}[!hbt]
    \centering
    \resizebox{\textwidth}{!}{%
    \begin{tabular}{cc|ccc|cccc}
        \hline
        Frame ID & Child Frame ID & $tx[\unit{\meter}]$ & $ty[\unit{\meter}]$ & $tz[\unit{\meter}]$ & $qx$ & $qy$ & $qz$ & $qw$ \\ \hline
        base\_link & sensor\_box\_link & $2.065$ & $0$ & $1.105$ & $0$ & $0$ & $0$ & $1$ \\
        sensor\_box\_link & reach\_1 & $0.04$ & $0.4$ & $0.263$ & $0$ & $0$ & $0$ & $1$ \\
        sensor\_box\_link & reach\_2 & $0.04$ & $-0.4$ & $0.263$ & $0$ & $0$ & $0$ & $1$ \\
        sensor\_box\_link & reach\_3 & $0.3825$ & $0$ & $0.239$ & $0$ & $0$ & $0$ & $1$ \\
        sensor\_box\_link & realsense\_bottom\_screw\_frame & $0.312269$ & $0$ & $0.051969$ & $0$ & $0.1564345$ & $0$ & $0.9876883$ \\
        realsense\_bottom\_screw\_frame & realsense\_infra1\_optical\_frame & $-0.011$ & $0.018$ & $0.013$ & $0.5$ & $-0.5$ & $0.5$ & $-0.5$ \\
        realsense\_infra1\_optical\_frame & realsense\_infra2\_optical\_frame & $0.050240$ & $0.000032$ & $0.000181$ & $0.0000482$ & $-0.0006391$ & $-0.0000144$ & $0.99999999$ \\
        realsense\_infra2\_optical\_frame & realsense\_color\_optical\_frame & $-0.06363861$ & $-0.00066263$ & $-0.00010325$ & $-0.00170204$ & $-0.00387651$ & $0.00520012$ & $0.99997752$ \\
        realsense\_infra1\_optical\_frame & realsense\_imu\_optical\_frame & $0.004398$ & $-0.019334$ & $-0.027490$ & $-0.002505$ & $0.000293$ & $-0.001948$ & $0.999994$ \\
        realsense\_imu\_optical\_frame & reach\_1\_imu & $-0.03328264$ & $-0.09014729$ & $-0.12951963$ & $-0.41522836$ & $0.42277546$ & $0.57759602$ & $0.56145272$ \\
        realsense\_imu\_optical\_frame & reach\_2\_imu & $0.05413669$ & $-0.05020996$ & $-0.1210384$ & $-0.4082114$ & $-0.5625849$ & $0.43126286$ & $0.57521651$ \\
        realsense\_imu\_optical\_frame & reach\_3\_imu & $-0.02761944$ & $-0.04919555$ & $-0.12547842$ & $-0.10502215$ & $-0.1182785$ & $0.6993272$ & $-0.69708107$ \\
        \hline
    \end{tabular}
    }
    \caption{Rigid transformation between coordinate frames.}
    \label{tab:transformations}
\end{table*}

\section{Appendix C: Odometry} \label{appendix:odometry}
We get the raw data from the wheel motor's microcontroller through serial communication.
These messages are recorded and made available in the dataset.
The serial message has the form:
$((<status>:[<motor nr>, <rpm>, <pulses>, <duty>, <current>]_{i:1\dots4}, <angle>, <backward>))$.
Where $status$ indicates whether the robot is in manual or automatic mode (all the sequences in this dataset are manual mode), $motor nr$ is an identifier of the wheel motor, $rpm$ is the linear velocity of the wheel and $pulses$ number of pulses detected by the sensors of the motors since the system is started.

\section{Appendix D: Precision of PPS-Based Time Synchronization} \label{appendix:chrony_precision}
``Chrony''~\footnote{\url{https://chrony-project.org/}}, the service we use to perform PPS-based synchronization, provides precise offset measurements that allow us to evaluate synchronization performance. In this appendix, we analyze the offset variations over each sequence to assess the accuracy and stability of time synchronization. The term offset refers to the difference between the system current time and a reference time source measured by Chrony and considered to gradually converge the system clock with the reference. In this case, the source is a GNSS receiver that, after convergence, achieves the precision of an atomic clock from satellites.

Figure~\ref{fig:chrony_offsets_gps}
illustrates the distribution of offset values for each sequence. Samples with values exceeding \num{1.5} times the interquartile range (IQR) beyond the quartiles are shown separately, as is typically done in a box plot. It should be mentioned that these outliers, marked as red crosses, are predominantly observed in the earlier intervals, suggesting a short period of instability (less than \SI{30}{\second}) before the system reaches synchronization.

Despite the presence of early outliers, the computed offsets for each sequence remain very small throughout the sequence, the mean being less than 0.32 milliseconds. This indicates that the synchronization process is performing well, considering that the fastest sensor operates at 200Hz producing measurements every 5 milliseconds. Offset mean values shown confirm that the system clock remains closely aligned with the reference GNSS satellite time effectively minimizes timing discrepancies, ensuring reliable and accurate timekeeping and stamping.
\begin{figure*}[!htbp]
    \centering
    \includegraphics[width=\linewidth]{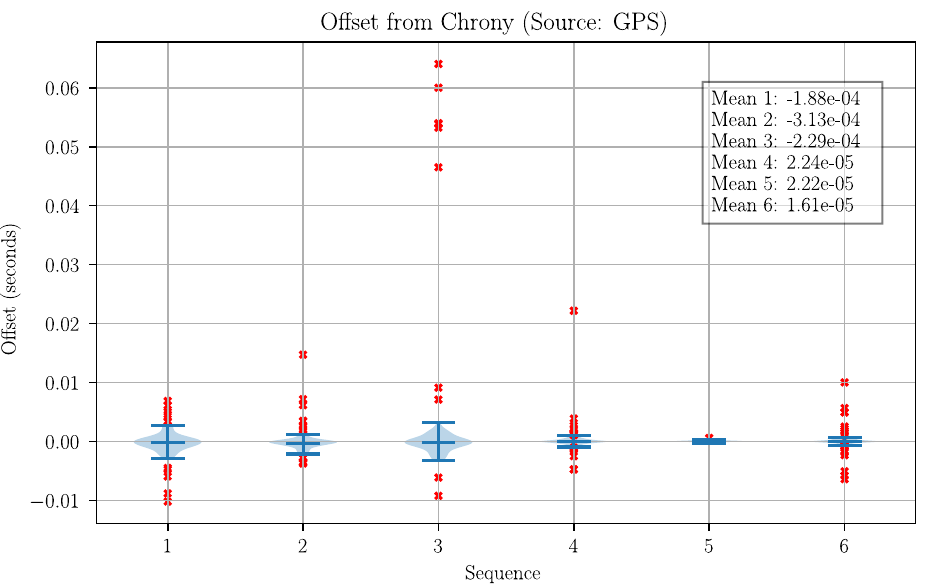}
    \caption{Distribution of offset values for each sequence in the PPS-based synchronization process. Outliers are marked as red crosses.}
    \label{fig:chrony_offsets_gps}
\end{figure*}

\section{Appendix E: File Structure} \label{appendix:file_structure}
We release the rosbags for the 6 recorded sequences, as well as some additional rosbags with postprocessed or independently recorded data, that can be easily merged or played synchronously as the timestamps line up.

\begin{itemize}
    \item \verb|2023_12_22_13_14_16.compressed.bag|: compressed rosbag for the first sequence of the first day (sequence \#1),
    \item \verb|2023_12_22_14_29_43.compressed.bag|: compressed rosbag for the second sequence of the first day (sequence \#2),
    \item \verb|2023_12_22_16_31_08.compressed.bag|: compressed rosbag for the third sequence of the first day (sequence \#3),
    \item \verb|2023_12_26_13_39_43.compressed.bag|: compressed rosbag for the first sequence of the second day (sequence \#4),
    \item \verb|2023_12_26_15_10_15.compressed.bag|: compressed rosbag for the second sequence of the second day (sequence \#5),
    \item \verb|2023_12_26_15_48_38.compressed.bag|: compressed rosbag for the third sequence of the second day (sequence \#6),
    \item \verb|*_conventional_gnss.bag|: rosbags with conventional GNSS data (no post-processing) for the corresponding sequence,
    \item \verb|*_ppk_gnss.bag|: rosbags with post-processed (PPK) GNSS data for the corresponding sequence,
    \item \verb|*_pgt.bag|: rosbags with post-processed ground-truth information as described in Section~\ref{sec:dataset} for the corresponding sequence
    .
\end{itemize}

These rosbags contain collision-free names for the different topics, making running or merging multiple rosbags hassle free.

Scripts are provided in the linked GitHub repository that allow extracting any and all rosbags into a common file directory structure, of images and \textit{comma-separated-value} files, and converting between common file representations where possible.

We also provide the rosbags recording for the calibration of the different sensors and the files resulting from running such calibrations:

\begin{itemize}
    \item \verb|reach_imus_bias.bag|: rosbag with a \SI{3}{\hour} recording of the data produced by the IMU units of the Reach sensors when stationary, useful for calculating the biases,
    \item \verb|reach_imu_biases.tar.gz|: compressed files resulting from running the Allan Variance ROS Compatible Tool\footnote{\url{https://github.com/ori-drs/allan\_variance\_ros}} on the previously mentioned rosbag,
    \item \verb|realsense_imus_bias.bag|: rosbag with a \SI{3}{\hour} recording of the data produced by the IMU unit of the RealSense D435i sensor when stationary, useful for calculating its bias,
    \item \verb|reach_imu_biases.tar.gz|: compressed files resulting from running the Allan Variance ROS Compatible Tool on the previously mentioned rosbag,
    \item \verb|allan_variance_config.tar.gz|: compressed files used for configuration of the Allan Variance ROS Compatible Tool,
    \item \verb|aprilgrid_calibration.bag|: rosbag with a calibration routine recorded with an aprilgrid board and making sure to excite all axes, useful for calibrating with Kalibr,
    \item \verb|kalibr_ir_color.tar.gz|: compressed files resulting from running the Kalibr Visual-Inertial Calibration Toolbox\footnote{\url{https://github.com/ethz-asl/kalibr}} on the previously mentioned rosbag. This calibration was performed using the IR stereo left camera and the color camera of the RealSense D435i only,
    \item \verb|kalibr_ir_stereo.tar.gz|: compressed files resulting from running the Kalibr Visual-Inertial Calibration Toolbox on the previously mentioned rosbag, using only the IR stereo left and right cameras of the RealSense D435i,
    \item \verb|kalibr_ir_stereo_imu.tar.gz|: compressed files resulting from running the Kalibr Visual-Inertial Calibration Toolbox on the previously mentioned rosbag. This calibration was performed using the IR stereo left and right cameras, along with the IMU of the RealSense D435i only,
    \item \verb|kalibr_ir_stereo_imu_imus.tar.gz|: compressed files resulting from running the Kalibr Visual-Inertial Calibration Toolbox on the previously mentioned rosbag.  This calibration was performed using the IR stereo left and right cameras of the RealSense D435i, its IMU, as well as the IMUs of the Reach sensors,
    \item \verb|kalibr_config.tar.gz|: compressed files used for configuration of the Kalibr Visual-Inertial Calibration Toolbox,
    \item \verb|magnetic_compass.tar.gz|: compressed files containing information about the magnetic compass, including magnetic declination (for different magnetic models) and calibrations generated from the trajectory sequences.
\end{itemize}

\subsection{Rosbag Topics}
In Table~\ref{tab:rosbag_topics} we list the different topics contained in the rosbags, their message types and a short description.

\begin{table*}[!htbp]
    \centering
    \resizebox{\textwidth}{!}{%
    \begin{tabular}{llc}
    \multicolumn{1}{c}{\textbf{Topic Name}} & \multicolumn{1}{c}{\textbf{Topic Type}} & \textbf{Topic Description}                                                                                      \\ \hline
    /distance                               & wheel\_odometry/Distances               &   Raw data from the wheel motor’s microcontroller                          \\
    /odom                                   & nav\_msgs/Odometry                      & \begin{tabular}[c]{@{}c@{}}Wheel odometry,\\  linear and angular velocity\end{tabular}                             \\
    /reach\_*/fix                           & sensor\_msgs/NavSatFix                  & \begin{tabular}[c]{@{}c@{}}Position data from the given\\ Real-Time Kinematic GNSS (RTK)\end{tabular}           \\
    /reach\_*/vel                           & geometry\_msgs/TwistStamped             & \begin{tabular}[c]{@{}c@{}}Velocity data provided by the\\ given GNSS module\end{tabular}                       \\
    /reach\_*/imu                           & sensor\_msgs/Imu                        & \begin{tabular}[c]{@{}c@{}}Inertial measurement unit data\\ from the given GNSS module\end{tabular}             \\
    /reach\_*/imu/mag                       & sensor\_msgs/MagneticField              & \begin{tabular}[c]{@{}c@{}}Magnetometer data from the \\ given GNSS module\end{tabular}                         \\
    /reach\_*/imu/driver\_t                 & sensor\_msgs/TimeReference              & \begin{tabular}[c]{@{}c@{}}Timestamps generated by the internal hardware \\ driver corresponding to the IMU measurements\end{tabular}                                  \\
    /reach\_*/imu/system\_t                 & sensor\_msgs/TimeReference              & \begin{tabular}[c]{@{}c@{}}Timestamps registered by the Emlid Reach module \\ system clock corresponding to the IMU measurements\end{tabular}                \\
    /reach\_*/time\_reference               & sensor\_msgs/TimeReference              & Timestamps from NMEA sentence of GNSS solution                 \\
    /realsense/infra1/camera\_info          & sensor\_msgs/CameraInfo                 & \begin{tabular}[c]{@{}c@{}}IR camera parameters published\\ by the RealSense ROS1 driver\end{tabular}           \\
    /realsense/infra1/image\_rect\_raw      & sensor\_msgs/Image                      & \begin{tabular}[c]{@{}c@{}}IR camera image published by\\ the RealSense ROS1 driver\end{tabular}                \\
    /realsense/infra2/camera\_info          & sensor\_msgs/CameraInfo                 & \begin{tabular}[c]{@{}c@{}}IR camera parameters published\\ by the RealSense ROS1 driver\end{tabular}           \\
    /realsense/infra2/image\_rect\_raw      & sensor\_msgs/Image                      & \begin{tabular}[c]{@{}c@{}}IR camera image published by\\ the RealSense ROS1 driver\end{tabular}                \\
    /realsense/color/camera\_info           & sensor\_msgs/CameraInfo                 & \begin{tabular}[c]{@{}c@{}}Color camera parameters published\\ by the RealSense ROS1 driver\end{tabular}        \\
    /realsense/color/image\_raw             & sensor\_msgs/Image                      & \begin{tabular}[c]{@{}c@{}}Color camera image published by\\ the RealSense ROS1 driver\end{tabular}             \\
    /realsense/depth/camera\_info           & sensor\_msgs/CameraInfo                 & \begin{tabular}[c]{@{}c@{}}Depth camera parameters published\\  by the RealSense ROS1 driver\end{tabular}       \\
    /realsense/depth/image\_rect\_raw       & sensor\_msgs/Image                      & \begin{tabular}[c]{@{}c@{}}IR depth image published by\\ the RealSense ROS1 driver\end{tabular}                 \\
    /realsense/accel/imu\_info              & realsense2\_camera/IMUInfo              & \begin{tabular}[c]{@{}c@{}}Accelerometer parameters published\\ by the RealSense ROS1 driver\end{tabular}       \\
    /realsense/gyro/imu\_info               & realsense2\_camera/IMUInfo              & \begin{tabular}[c]{@{}c@{}}Gyroscope parameters published\\ by the RealSense ROS1 driver\end{tabular}                                                                                                                \\
    /realsense/imu                          & sensor\_msgs/Imu                        & \begin{tabular}[c]{@{}c@{}}Inertial measurement unit data published\\ by the RealSense ROS1 driver\end{tabular} \\
    /tf                                     & tf2\_msgs/TFMessage                     &  Dynamic transformations between coordinate frames                                                               \\
    /tf\_static                             & tf2\_msgs/TFMessage                     & Static transformations between coordinate frames                                                                 \\
    /reach\_*/gps/fix                       & sensor\_msgs/NavSatFix                  & Single Point Positioning GNSS (SPP)                                                                                 \\
    /reach\_*/gps/vel                       & geometry\_msgs/TwistStamped             & Velocity data from SPP GNSS                                                                                         \\
    /reach\_*/ppk/fix                       & sensor\_msgs/NavSatFix                  & Post-Processed Kinematic GNSS (PPK)                                                                                   \\
    /reach\_*/ppk/vel                       & geometry\_msgs/TwistStamped             & Velocity data from PPK GNSS                                                                                        
    \end{tabular}%
    }
    \caption{Recorded or post-processed topic names, their associated type, and a description of the data it contains. An asterisk implies that there are multiple topics similarly named with the same type and description.}
    \label{tab:rosbag_topics}
\end{table*}

\section{Appendix F: Supplementary Dataset} \label{appendix:zavalla_2021}
As a preliminary step in recording this dataset, we collected a large amount of agricultural data in previous field trips. This data can be of great utility for evaluating SLAM algorithms, sensor fusion, path planning, navigation and control. For this reason we have decided to release the data captured in these field trips. In this appendix, we provide detailed information about this supplementary dataset, highlighting the differences between it and the core data presented in this article.

\subsection{The Robot}
The main difference with respect to The Rosario Dataset 2023 is the sensors mounted onboard the robot. Figure~\ref{fig:supp_dataset_robot} shows the weed removing robot at the time of recording the supplementary dataset. As can be seen, the sensor layout differs here, as there is no sensor box. To capture the data, the sensors are connected to a
NVIDIA Jetson TX2. The rest of the robotic platform is the same as described in Section~\ref{sec:platform}.

\begin{figure}[!htbp]
    \centering
    \includegraphics[width=\linewidth]{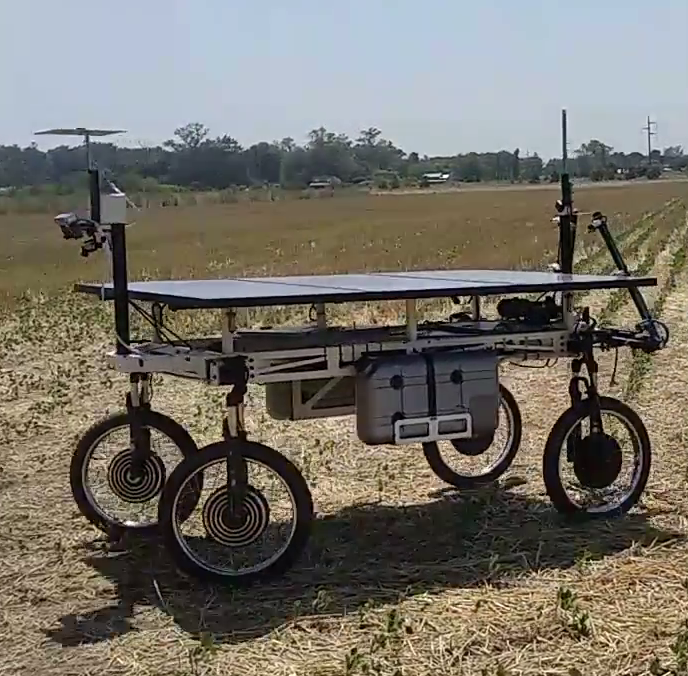}
    \caption{Weed removing robot at the time of recording the supplementary dataset.}
    \label{fig:supp_dataset_robot}
\end{figure}

\begin{table*}[!htbp]
    \centering
    \begin{tabular}{cccc}
    \hline
    \textbf{Name} &
      \textbf{Sensor} &
      \textbf{\begin{tabular}[c]{@{}c@{}}Resolution /\\ Range\end{tabular}} &
      \textbf{Acquisition Rate} \\ \hline
  \begin{tabular}[c]{@{}c@{}}Zed Camera\end{tabular} &
  Stereo Color Camera &
  \begin{tabular}[c]{@{}c@{}}1280 × 720 and 672 x 376\\ \SI{90}{\degree} × \SI{60}{\degree}\end{tabular} &
  \SI{15}{\hertz} \\ \hline
    \multirow{5}{*}{Emlid Reach M1}     & SPP     & ± \SI{2.5}{\meter}                                                     & \SI{5}{\hertz}   \\ \cline{2-4} 
                                        & RTK-GNSS & ±\SI{0.04}{\meter}                                                     & \SI{5}{\hertz}   \\ \cline{2-4} 
                                        & 6-DoF IMU      & \begin{tabular}[c]{@{}c@{}}± \num{8} g\\ ± \SI{1000}{\deg\per\second}\end{tabular}    & \SI{200}{\hertz} \\ \cline{2-4}
                                        & Magnetometer & - & \SI{40}{\hertz} \\ \hline
    \begin{tabular}[c]{@{}c@{}}E-Bike Wheel\\ Odometer\end{tabular} &
      Hall-Effect Odometry &
      ± \SI{7.5}{\degree} &
      \SI{10}{\hertz} \\ \hline
    \begin{tabular}[c]{@{}c@{}}OMRON \\ E6CP-A\end{tabular} &
    \begin{tabular}[c]{@{}c@{}}8-bit Absolute Encoder\end{tabular} &
    \begin{tabular}[c]{@{}c@{}}± \SI{1.4}{\degree}\\ \SI{92}{\degree}\end{tabular} &
      \SI{10}{\hertz} \\ \hline
    \end{tabular}
    \caption{List of sensors mounted on the platform at the time of recording the supplementary dataset.}
    \label{tab:supp_dataset_sensor_list}
\end{table*}

\subsection{Data Collection Systems Overview}

The robot navigation and data recording stack was implemented on the ROS framework. For each sensor there is a ROS node capable of taking data from the sensor and converting it to a ROS message. If available, we use the official implementation of the node provided by the sensor manufacturer. All the messages coming from the sensors are stored in a single rosbag for each trajectory performed except for the images. Due to the size of the disk space occupied by the images, we chose to use the SVO format provided by the Zed camera and created by StereoLabs. This format compresses the images captured by the camera allowing them to be stored with significantly less disk space. Tools that allow conversion from SVO to rosbag are available.

\subsection{Dataset}

The tests were conducted on December 28th and 29th, 2021. Henceforth, we will refer to these working days as Day 1 and Day 2, respectively. Seven recording sessions were conducted between the two days, totalling seven data sequences. The sequences are numbered in chronological order, with sequence 1 being the first sequence recorded on Day 1 and sequence 7 the last sequence captured on Day 2. For all trajectories, data from the stereo cameras, IMU, magnetometer, GNSS-RTK, velocity from GNSS, SPP from GNSS, and wheel odometry are captured and recorded. Additionally, raw data provided by the GNSS receiver modules are stored. This data is processed offline using PPK to obtain a higher GNSS position accuracy than that achieved with RTK. 

On Day 1 of the fieldwork, two distinct trajectories were performed in the soybean field, resulting in Sequences 1 and 2. At the start of Sequence 1, calibration routines were conducted, involving the robot rotating in circles in both directions, as described in Section~\ref{sec:dataset} for the 2023 dataset. 
On Day 2 of fieldwork, five data sequences were collected (Sequences 3, 4, 5, 6, and 7). Similar to Sequence 1, Sequences 3 and 4 begin with calibration routines.

\subsection{Synchronization}
Each sensor measurement includes a timestamp. In this setup of the weed removing robot, the sensors were not hardware-synchronized. SLAM algorithms generally assume that input data is synchronized on a single clock, so this lack of synchronization poses a challenge when using our data with SLAM algorithms.

To implement software-based synchronization, we identify moments in the sequence where the robot remains stationary for at least $n$ seconds and then begins moving, detecting these timestamps with each sensor involved. We then select one sensor as the reference and apply an offset to each of the other sensors to align them with the reference sensor. Although the absolute timing might not be precise, our primary goal is to correct the relative timing between sensors. The IMU is chosen as the reference sensor because its high message frequency provides greater timing granularity.

\subsection{Ground-Truth}
Having a high-precision ground-truth is essential for evaluating SLAM systems, as it allows for comparing SLAM estimates against a reference trajectory. We planned to estimate PPK as outlined in Section~\ref{sec:ground_truth}.
However, an issue with the base station in one recording session on Day 1 caused it to shut down and stop transmitting corrections.

With the aim of addressing this issue, we utilized a station from the RAMSAC network for receiving the corrections. RAMSAC is a network of GNSS reference stations across Argentina, managed by the Argentine National Geographic Institute (IGN), providing a precise and uniform geodetic reference framework nationwide. This network, with stations distributed across Argentina, supports real-time monitoring and correction of GNSS data and is suitable for both RTK and PPK applications. Each station in the network is equipped with high-precision GNSS receivers that record data continuously, ensuring a constant stream of information.

The nearest RAMSAC station to our fieldwork site is the UNRO station, located at National University of Rosario, approximately 25 kilometers from the soybean field. Although this distance is at the optimal range limit (\SI{20}{\kilo\meter} for single-frequency receivers, according to RAMSAC website), the raw GNSS data from this station was used only to calculate PPK for the few minutes of recording when our local base station stopped transmitting. For all other data, PPK was processed using measurements from the local base station whenever available. 

For orientation ground-truth, we rely on the magnetometer and IMU measurements. Specifically, we perform a 2D magnetometer calibration using data recorded during the calibration routines. We then correct the magnetic measurements and estimate the global orientation using a Madgwick filter \cite{didomenico2023orientation, didomenico2024thesis}.

\end{sm}

\end{document}